\documentclass[runningheads]{llncs}
\usepackage{graphicx}
\usepackage{amsmath,amssymb} 
\usepackage{color}
\usepackage{float}
\usepackage[width=122mm,left=12mm,paperwidth=146mm,height=193mm,top=12mm,paperheight=217mm]{geometry}
\begin{document}

\pagestyle{headings}
\mainmatter

\title{Diverse feature visualizations reveal invariances \\in early layers of deep neural networks} 

\titlerunning{Diverse feature visualizations of early layers of CNNs}

\authorrunning{SA Cadena, MA Weis, LA Gatys. M Bethge, AS Ecker}

\author{Santiago A. Cadena \and Marissa A. Weis \and Leon A. Gatys \and\\
Matthias Bethge \and Alexander S. Ecker}

\institute{Centre for Integrative Neuroscience and Institute for Theoretical Physics\\
Bernstein Center for Computational Neuroscience\\
University of T\"ubingen, Germany \\
\email{\{first.last\}@bethgelab.org}}

\maketitle

\begin{abstract}
Visualizing features in deep neural networks (DNNs) can help understanding their computations.
Many previous studies aimed to visualize the selectivity of individual units by finding meaningful images that maximize their activation.
However, comparably little attention has been paid to visualizing to what image transformations units in DNNs are invariant.
Here we propose a method to discover invariances in the responses of hidden layer units of deep neural networks.
Our approach is based on simultaneously searching for a batch of images that strongly activate a unit while at the same time being as distinct from each other as possible. We find that even early convolutional layers in VGG-19 exhibit various forms of response invariance: near-perfect phase invariance in some units and invariance to local diffeomorphic transformations in others. At the same time, we uncover representational differences with ResNet-50 in its corresponding layers. We conclude that invariance transformations are a major computational component learned by DNNs and we provide a systematic method to study them.
\keywords{Feature visualization, invariance, phase invariance, deep neural networks, early visual system.}
\end{abstract}

\section{Introduction}

As deep neural networks have gained popularity in many scientific disciplines and technological applications, there is a growing interest in understanding the representations they learn and the computations they perform. One approach towards achieving such understanding is to visualize the features that activate the neurons in a network. There is a growing body of work that seeks to visualize features by synthesizing images which maximally drive hidden layer units. While this approach can give us a rough intuition about a unit's selectivity, it provides only a very incomplete picture of its computation. In addition to characterizing feature detectors by the stimulus that elicits the largest response, it is important to identify the nuisance parameters to which the neuron is invariant. As hidden layers build up response invariances gradually with depth, it is not the \emph{image} that most strongly drives a unit that is the most telling about this unit's function, but instead the \emph{set of images} that elicit a strong response. While some previous work has visualized multiple `facets' of neurons' selectivity, these efforts focused mostly on the highest layers of the network and relied on initialization or random sampling strategies to create multiple images for each unit. However, as we show in the present paper, these approaches underestimate the true diversity of the selectivity of even relatively low-level units. Additionally, these approaches have not offered insights about how the representations of different networks trained on the same task compare. Our contributions are the following:
\begin{enumerate}
    \item Motivated by the phase invariance of complex cells in the early visual system of the brain, we show why visualizing invariance is as important as visualizing selectivity for understanding the computations of even low-level units.
    \item We develop a non-parametric approach to map the manifold of highly-activating inputs as exhaustively as possible.
    \item We show that even relatively low-level units exhibit a remarkable degree of invariance in VGG-19 \cite{S2}, which is not revealed by finding highly activating stimuli from multiple optimization runs with random initializations.
    \item We find that in low to intermediate layers  of VGG-19, at least two types of invariances emerge: tolerance to local diffeomorphic transformations tuned to specific features, and phase invariance, where units respond well to periodic texture patterns and are insensitive to their phase. We additionally offer a way to quantify these invariances. 
    \item In contrast, we find that low to intermediate layers of a network with skip connections (ResNet-50 \cite{he2016deep}) that was trained on the same task as VGG-19 exhibit far less phase invariance, revealing representational differences between these two networks. 
    \item We showcase our visualization approach on a CNN trained to predict responses to natural images in primary visual cortex of the primate brain. 
\end{enumerate}
We provide the code to replicate our results. \footnote{https://github.com/sacadena/diverse\_feature\_vis}

\section{Related work}

One way to identify selectivity of hidden units is to look for image patches in the dataset that drive them maximally \cite{E1,Z1}. These image patches can sometimes hint at a unit's selectivity, but it can be difficult to identify their common features. Optimization-based techniques have proven more useful for feature visualization: a common approach is to search for pre-images that drive individual neurons maximally via gradient ascent~\cite{E1}. Most previous work focused on deep layers, where finding natural-looking pre-images is challenging. For example, the activation objective leads to adversarial-like patterns \cite{N3,Sz1}. As a consequence, much of the follow-up work focused on developing regularization techniques to obtain more natural pre-images, including penalties on high-frequency noise \cite{M2,N3} or the distance between the generated visualizations and natural images patches \cite{W1}, or performing gradient descent in the feature space of a deep generator network \cite{N4}.
 
Goodfellow et al. \cite{G1} were the first (to our knowledge) to study invariances in deep networks. Their approach allows to quantify how invariant a unit is to known transformations such as translation, (3D-) rotation or scaling, but it does not allow to discover these transformations if they are unknown in advance. 

Recent work proposes visualizing multiple `facets' of the neuron's selectivity by obtaining multiple images from different random initializations \cite{M1}, using a diverse set of highly activating images as initializations \cite{N2}, or using a generative image model to sample highly-activating images \cite{N1}. 

These methods do not explicitly specify an objective to produce a diverse set of images. In contrast, we optimize a batch of images to drive the neuron of interest strongly while simultaneously being as distinct from each other as possible. Recent concurrent work \cite{O1} introduces a similar idea, albeit with a different loss function based on texture representations \cite{Ga2,Ga1}.

\section{Discovering invariances}

\subsection{Motivation: simple and complex cells}

\begin{figure}[t]
\centering
\includegraphics[width=\linewidth]{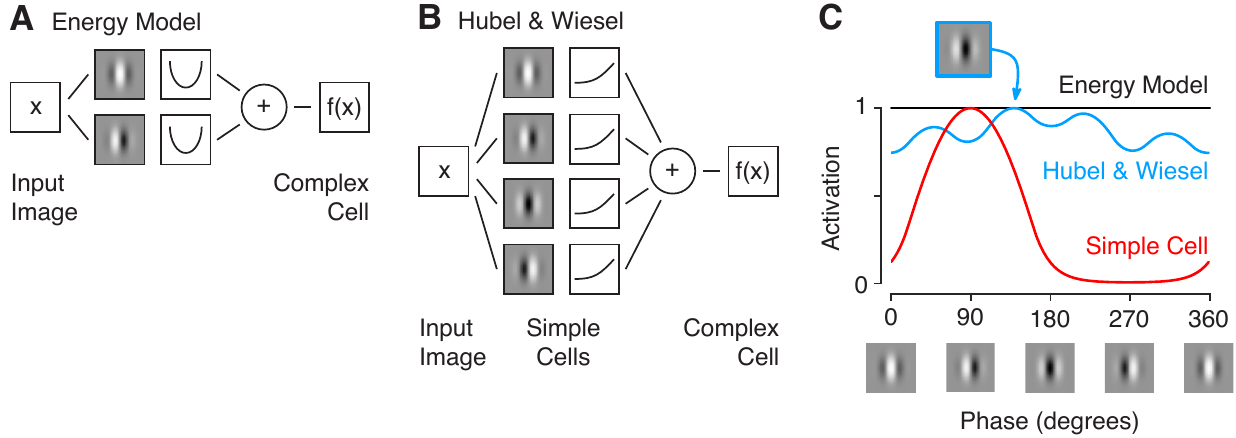}
\caption{
Simple and complex cells, phase invariance.
{\bf A.} Energy model of complex cell.
{\bf B.} Hubel \& Wiesel model of complex cell.
{\bf C.} Neural response as a function of phase of Gabor stimulus with optimal orientation and spatial frequency.
}
\label{fig:complex}
\end{figure}

We illustrate our point by considering a toy example well known from early vision in the brain (Fig.~\ref{fig:complex}): simple and complex cells~\cite{Hu1}, which are found in the primary visual cortex, an early stage of visual processing in the mammalian brain. Simple cells can be approximated well by a linear filter followed by a thresholding nonlinearity (e.g. ReLU). The linear filter usually resembles a Gabor filter. Complex cells are, like simple cells, selective for a specific orientation and spatial frequency. However, unlike simple cells they respond to Gabor patches of arbitrary phases -- they are phase-invariant. The standard model for this phase invariance is the so-called energy model (Fig.~\ref{fig:complex}A, \cite{A1}), which sums over the squared responses of two Gabor filters phase-shifted by 90$^\circ$ (Fig.~\ref{fig:complex}C, black). This energy model has also been used to study rotation, scaling and more general invariances in the context of unsupervised representation learning \cite{berkes2005slow,bethge2007unsupervised,lies2014slowness}

An alternative formulation was originally proposed by Hubel \& Wiesel, who discovered complex cells in the 1960ies in the primary visual cortex of cats \cite{Hu1}. Their model suggests that complex cells are the result of pooling over multiple simple cells with a range of phase preferences (Fig.~\ref{fig:complex}B). If the learned weights and phase preferences exhibit some variability, the resulting phase invariance is only approximate (Fig.~\ref{fig:complex}C, blue).

Now, consider what happens when we study simple and complex cells using activity maximization. For a simple cell, we will recover its selectivity. For a complex cell, however, all Gabor patches of optimal orientation and spatial frequency will elicit a high response, irrespective of their phase. In the case of the Energy Model, which is perfectly phase-invariant, we may obtain this set of optimal images by starting with random initializations. However, for an imperfect model more likely to occur in reality (e.\,g. Hubel \& Wiesel model, blue in Fig.~\ref{fig:complex}C), there is a unique maximum, which we will find despite the fact that activations are consistently above 80\% of the maximum for all phases. Thus, activity maximization will produce the same result for both simple and complex cells (a single Gabor patch), but this result will miss the key aspect of the complex cell's computation: its phase invariance.

\subsection{Mapping invariances}

\subsubsection{Objective.}

The idea behind our approach is to find a batch of images in which each image maximally drives a specific unit while the images are maximally different from one another. Starting with a batch of $n$ images $\{x_{1}, \cdots, x_{n}\}$, initialized as white noise, we \emph{maximize} the following objective using gradient ascent:

\begin{equation}\label{eq:obj}
L = \sum_{i=1}^n y_{ik}^{(l)} + \alpha \sum_{i=1}^n \log P(x_i) + \lambda \min_{i,j}d(x_i, x_j).
\end{equation}
Here, $y_{ik}^{(l)}$ is the output activation of unit $k$ in layer $l$ for the $i^\mathrm{th}$ image in the batch, $P(x_i)$ is the likelihood of the image under a generative model of natural images and $d(x_i, x_j)$ is a distance between two images, The likelihood and distance measures are specified below. Note that we set the image size to the receptive field size of units in the layer to be visualized, such that the outputs $y_{ik}^{(l)}$ are $1\times 1$ spatially and we can omit the indices over space. We constrain the norm of the synthesized images to be equal to half the average norm of natural images patches of the same size taken from the ImageNet dataset\footnote{Using half the average norm is a heuristic that we use because the synthesized images tend to be localized to the center of the patch.}, where we assume that zero in each color channel corresponds to the average value of this channel across the ImageNet training set. For visualization, we add this mean and clip the values between 0 and 255. Very few pixels fall outside this range.

The first and the second term in the objective are similar to previous work, encouraging the optimization to find natural images that strongly activate the unit. The third term forces all images in the batch to be as distinct as possible from all other images, since we penalize the minimum distance between any pair of images. This objective presents a trade-off: we allow for some degree of non-maximal responses if this allows us to increase the set of strongly activating pre-images substantially. 

It is important to use the minimum distance in the objective rather than the average. Maximizing the average distance does not necessarily lead to coverage of the invariant subspace. Consider the Energy Model: assuming we generate an even number of $n$ images, the optimal solution maximizing the average $L_2$ distance is to place all images at either of two distinct phases separated by $180^\circ$. Now we fail to generate a diverse set of images but the average distance is high ($90^\circ$). In contrast, the desired solution of images evenly separated by $360^\circ/n$ will give a smaller average distance for $n>4$ and can be obtained when maximizing the minimum distance.

It has also some advantages to consider a single unit within a feature map compared to considering the entire feature map. When maximizing the activation of the entire feature map, the resulting image will be shift-invariant by construction and properties such as phase invariance of individual units cannot be detected.

\subsubsection{Natural image prior.}
We use PixelCNN++ \cite{Sa1} as a natural image prior, as it allows directly evaluating and optimizing the likelihood of an image patch of arbitrary size. 
In a nutshell, PixelCNN++ improves upon PixelCNN \cite{van2016conditional} and earlier autoregressive models \cite{oord2016pixel,theis2015generative,theis2012mixtures} that attempt to capture the distribution of natural images by expressing the joint distribution of all pixels as the product of the distributions of individual pixels conditioned on a causal neighborhood. We use the model pre-trained on Cifar-100 provided by OpenAI\footnote{https://github.com/openai/pixel-cnn} which is state-of-the-art in terms of likelihood on natural images.

\subsubsection{Distance metric.}
To evaluate the distance between two images, we use a feature space given by the neural network to encourage diversity on perceptually interesting image properties. For an output unit $y_k$ in layer $l$, we compute the Euclidean distance in the feature space of the preceding convolutional layer:
\begin{equation}
d(x_i,x_j) = \Vert \textbf{y}_{i}^{(l-1)} - \textbf{y}_{j}^{(l-1)}\Vert_2;  \quad i \neq j 
\end{equation}
where $\textbf{y}_i^{(l-1)}$ and $\textbf{y}_j^{(l-1)}$ are vectors of activations in the preceding layer flattened over space and channels.

\subsubsection{Optimization.}
We optimize the objective defined in Eq.~(\ref{eq:obj}) using the Adam optimizer \cite{Ki1} with a learning rate of 0.1 until the objective converges (maximum of 1000 steps). Similar to Olah et al. \cite{O1}, we precondition the gradient to reduce the effect of high frequencies by dividing each frequency component by $\sqrt{f}$.

We manually set the hyperparameter $\alpha$, which controls the strength of the natural image prior, based on qualitative inspection of the resulting images in an exploratory experiment. We used $\alpha=0.0005$ for all experiments.

We sweep a range of values for $\lambda$ (0.02, 0.04, 0.08, ... 20.48) and for each unit pick the largest such $\lambda$ that the average activation level remains above a threshold. This threshold is 80\% of the maximum for the complex cell model and 90\% for VGG-19 and ResNet-50. See Fig.~\ref{fig:method}A and Fig.~\ref{fig:vgg_invariances} for a qualitative justification of these thresholds.

\subsection{Application to complex cell models}

\begin{figure}[t]
\centering
\includegraphics[width=83mm]{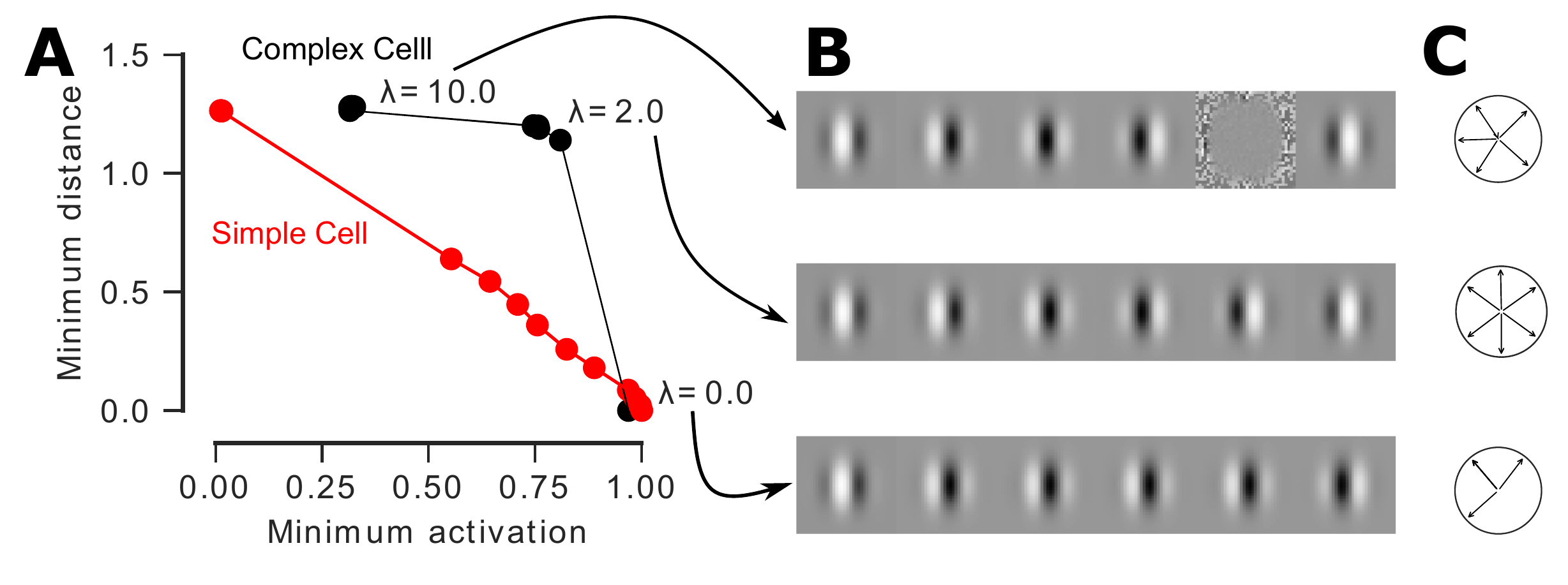}
\caption{Mapping invariances as a trade-off between diversity and maximizing activation.
{\bf A.} Trade-off between activation and image diversity. For a complex cell, images can be made quite diverse while keeping the activation level high. When $\lambda$ gets too large ($\lambda > 2$), there is a qualitative change.
{\bf B.} Set of images for three different $\lambda$.
{\bf C.} Distribution of phases of synthesized Gabor patches, showing that with the optimal $\lambda=2$ we get equally spaced images, i.e. cover the invariant subspace well.
}
\label{fig:method}
\end{figure}

Before applying our approach to a deep neural network, we verify that it works when the units are only approximately invariant to some transformation. To this end, we use the Hubel \& Wiesel model of a complex cell outlined above (Fig.~\ref{fig:complex}B), which does not produce perfect phase invariance, but still responds strongly to Gabor patches of all phases.

Indeed, our approach can visualize the entire invariant subspace spanning the full range of phases (Fig.~\ref{fig:method}). Without the diversity term ($\lambda=0$), the optimization tends to converge to the same pre-image (Fig.~\ref{fig:method}B). Four out of six solutions correspond to the globally most strongly driving image (see also Fig.~\ref{fig:complex}C, top). In contrast, with an appropriate choice of $\lambda$, the images distribute uniformly (Fig.~\ref{fig:method}B,\,C). 
If we increase $\lambda$ too much, however, the diversity penalty becomes too large and the optimization will converge to solutions including non-optimal images. Thus, to visualize the invariant subspace, we should pick the largest $\lambda$ that leads to only a small decrease in activation level. This point depends on how `clean' the invariance of the cell is. For the Hubel \& Wiesel model considered here, this drop in activation occurs when the average activation falls below 80\% of the maximum, which corresponds to the response range for images within the approximately invariant subspace (see Fig.~\ref{fig:complex}C, blue line).

Note that for the simple cell, which does not exhibit any such response invariance, the curve looks qualitatively different (Fig.~\ref{fig:method}A, red line). Thus, we can quantify response invariance of units in a DNN by computing the minimum distance between any two images in the batch at the optimal $\lambda$.

\section{Invariances in VGG-19}

We asked to what extent deep neural networks trained on large-scale object recognition (ImageNet \cite{R1}) exhibit response invariances in their convolutional layers. Previous work focused mostly on higher layers and did not find much invariance in low and intermediate layers. However, in neuroscience it is well-known that low- and mid-level neurons in the brain -- like complex cells -- can exhibit a substantial degree of response invariance. Moreover, there is evidence for a considerable degree of similarity between neural representations in DNNs trained on object recognition and the primate visual system \cite{Kr1,Gu1,C1,Ca1}. In particular, we have shown~\cite{C1} that the convolutional layers of VGG-19 \cite{S2} around layer conv3\_1 best predict neural activity in primary visual cortex, including that of many complex cells. Therefore we would expect that these layers in the VGG-19 network should also exhibit some degree of invariance to phase and potentially other transformations.

\subsection{Convolutional layers of VGG-19 exhibit response invariances}

\begin{figure}[t]
\centering
\includegraphics[width=110mm]{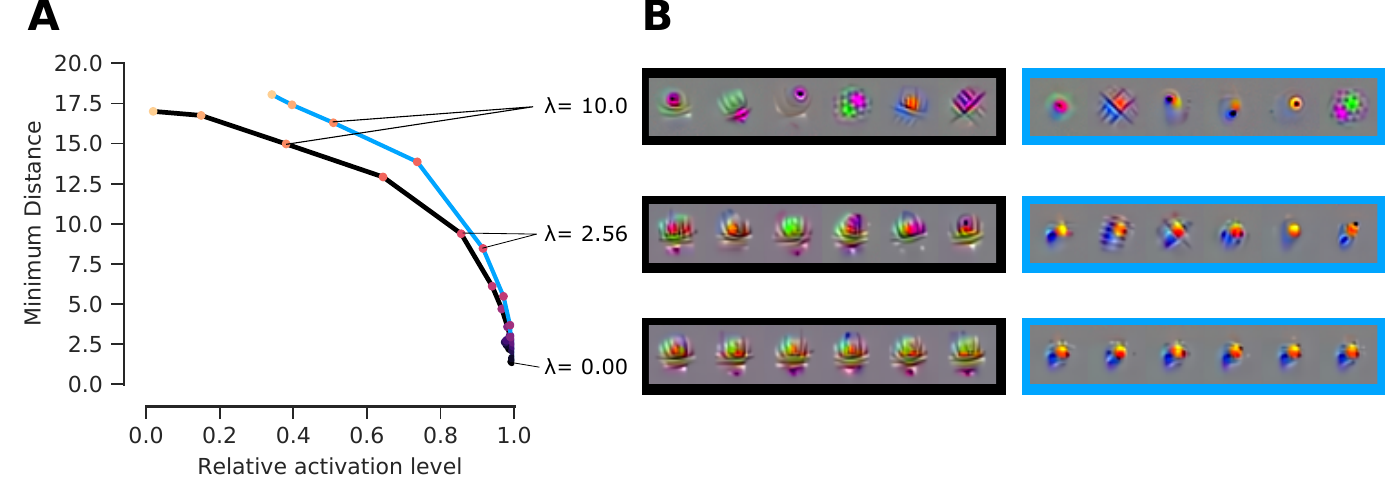}
\vspace{-6pt}
\caption{Invariant subspace of two example units in conv3\_2 (feature maps 9 and 26).
{\bf A.} Invariance/activation trade-off
{\bf B.} Pre-images obtained for different values of $\lambda$.
}
\label{fig:vgg_example}
\end{figure}

\begin{figure}[t!]
\begin{minipage}[c]{0.32\linewidth}
\includegraphics[width=\linewidth]{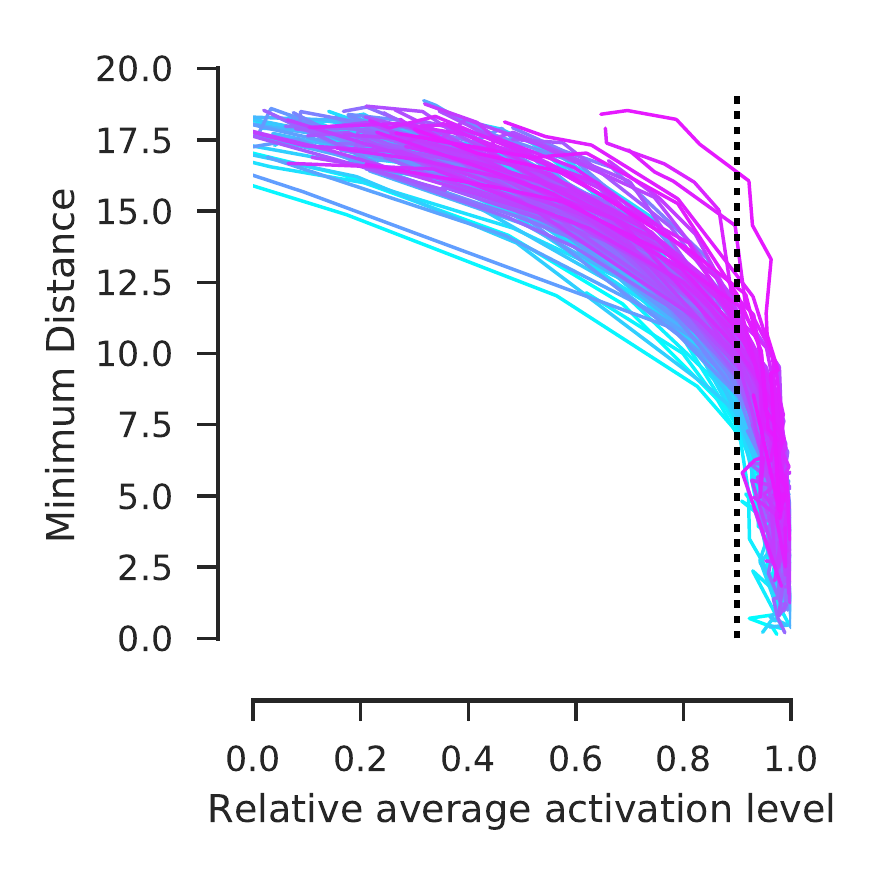}
\caption{Invariance/activity maximization trade-off for all units in layer conv3\_1. Based on visual inspection we deem 90\% an appropriate threshold and use the largest $\lambda$ such that the average activation remains above 90\% of the maximum activity.}
\label{fig:vgg_invariances}
\end{minipage}
\hfill
\begin{minipage}[c]{0.61\linewidth}
\includegraphics[width=0.9\linewidth]{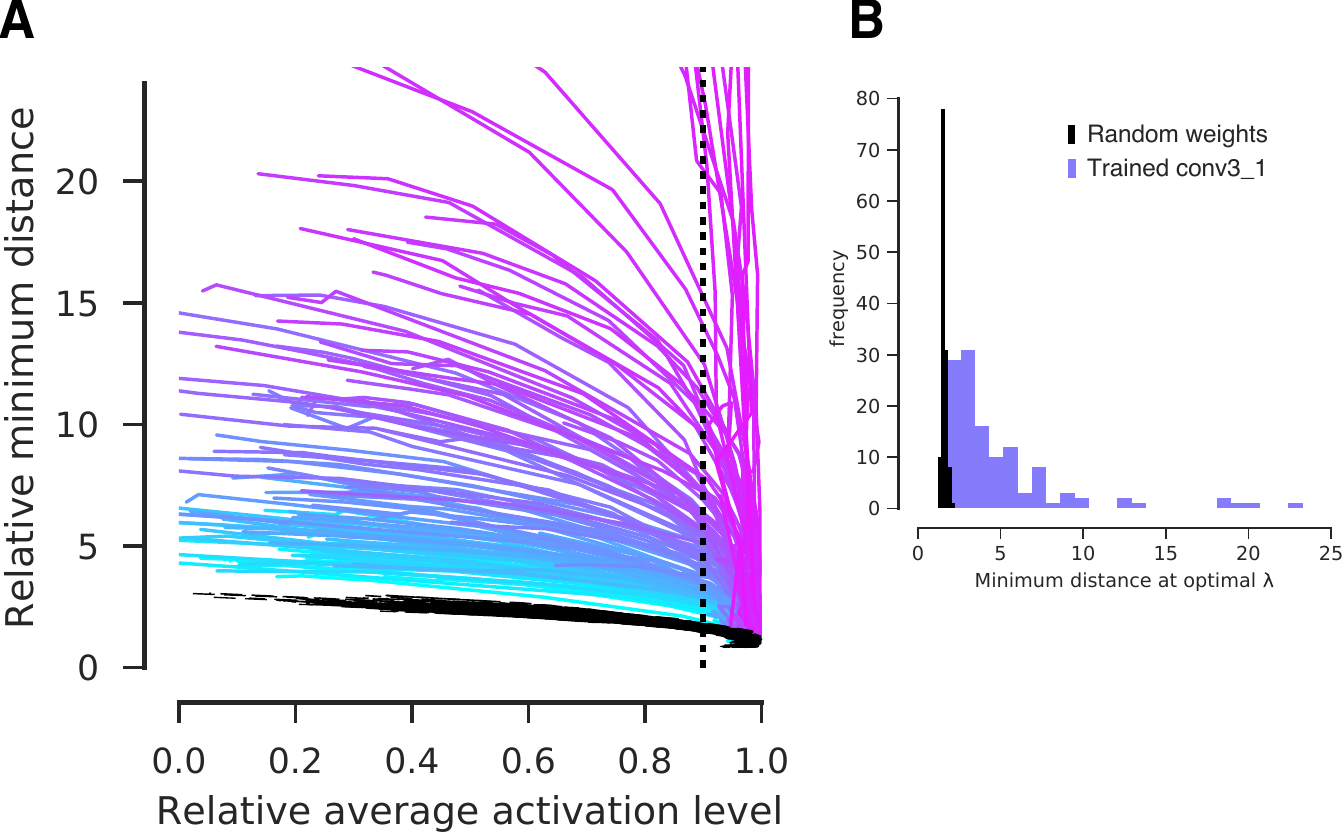}
\caption{VGG units are more invariant than expected from random weights.
{\bf A.} All 256 units in conv3\_1 (colored lines) are more invariant than units in a network with the same architecture but random weights (black lines).
{\bf B.} Histogram of the diversity terms for the optimal $\lambda$ relative to their value for $\lambda=0$ (black: random weights; purple: trained conv3\_1). This means that for the least invariant units we can increase the diversity of the images two-fold while maintaining the average activation above 90\% of the maximum obtained with $\lambda=0$.}
\label{fig:invariance_control}
\end{minipage}%
\end{figure}

We start by considering two example units from layer conv3\_2 (Fig.~\ref{fig:vgg_example}) of VGG-19. As in the complex cell example, we can increase the diversity of generated images quite substantially while maintaining a high activation level (Fig.~\ref{fig:vgg_example}A). Only when we increase $\lambda$ too much, the activation level drops substantially and the images start deteriorating (Fig.~\ref{fig:vgg_example}B, top row). Overall, the trade-off between image diversity and activation level looks qualitatively similar to the complex cell example above.

Moreover, the images generated with the optimal $\lambda$ look significantly more diverse than those obtained by random initialization at $\lambda=0$ (Fig.~\ref{fig:vgg_example}B, middle and bottom rows). Indeed, most units showed quite some degree of invariance: we can increase the image diversity considerably while maintaining activation levels above 90\% of the maximum (Fig.~\ref{fig:vgg_invariances} for conv3\_1; see Sect. 1 in the Supp. for additional convolutional layers). Below, we therefore use the largest such $\lambda$ that maintains the average activation level above 90\% of the maximum.

\subsection{Response invariances are a learned property of the network}

Is this invariance a learned property of the network or does it arise trivially from the network architecture? We repeated the analysis on a network with the same architecture as VGG-19 but random weights. To keep the two networks comparable, we normalized both the activations and the distances between images such that they are equal to one for $\lambda=0$. We found that units in the random network are substantially less invariant than those of VGG-19 (Fig.~\ref{fig:invariance_control}A), suggesting that the neurons' response invariance is indeed a learned property. Remarkably, by introducing the diversity term into the pre-image search, we could increase the minimum distance between any two images in a batch by a factor of at least two and up to 100-fold without `sacrificing' more than 10\% of the unit's activation level (Fig.~\ref{fig:invariance_control}B), a property that the random network does not exhibit.

\subsection{Types of invariance: texture vs. shape detectors}

\begin{figure}[t]
\centering
\includegraphics[width=90mm]{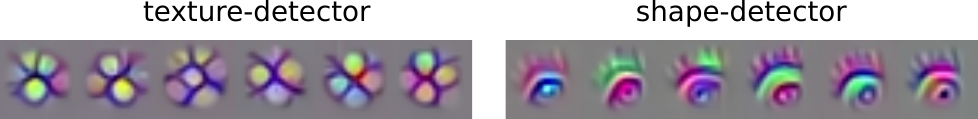}
\caption{Examples of invariant subspaces of texture-like and shape-like detectors of feature maps 13 (left) and 22 (right) in conv3\_1.}
\label{fig:two_examples}
\end{figure}

We now investigate the types of invariance learned by different units in the network. We start by considering two example units from layer conv3\_1 (Fig.~\ref{fig:two_examples}). The first unit responds to a dark grid on brighter background of arbitrary color. In addition to this selectivity, it appears to be entirely phase- and rotation-invariant: the location of the grid lines and their orientation is irrelevant for the unit's activation, but their general spatial scale and the foreground color are important. We refer to units that exhibit this property as \emph{texture} detectors.

The second unit, in contrast, detects a circular feature in the lower half of its receptive field. While it is sensitive to the location of this pattern within its receptive field, it exhibits a substantial degree of color and scale invariance: the contours have a sinusoidal cross-section whose local phase varies across images, such that by using linear combinations of multiple of these images one can obtain the circular pattern in various different sizes and color combinations. We refer to such units as \emph{shape} detectors: they are sensitive to location but allow for some degree of local diffeomorphic transformation.

\begin{figure*}[ht!]
\centering
\includegraphics[width=\linewidth]{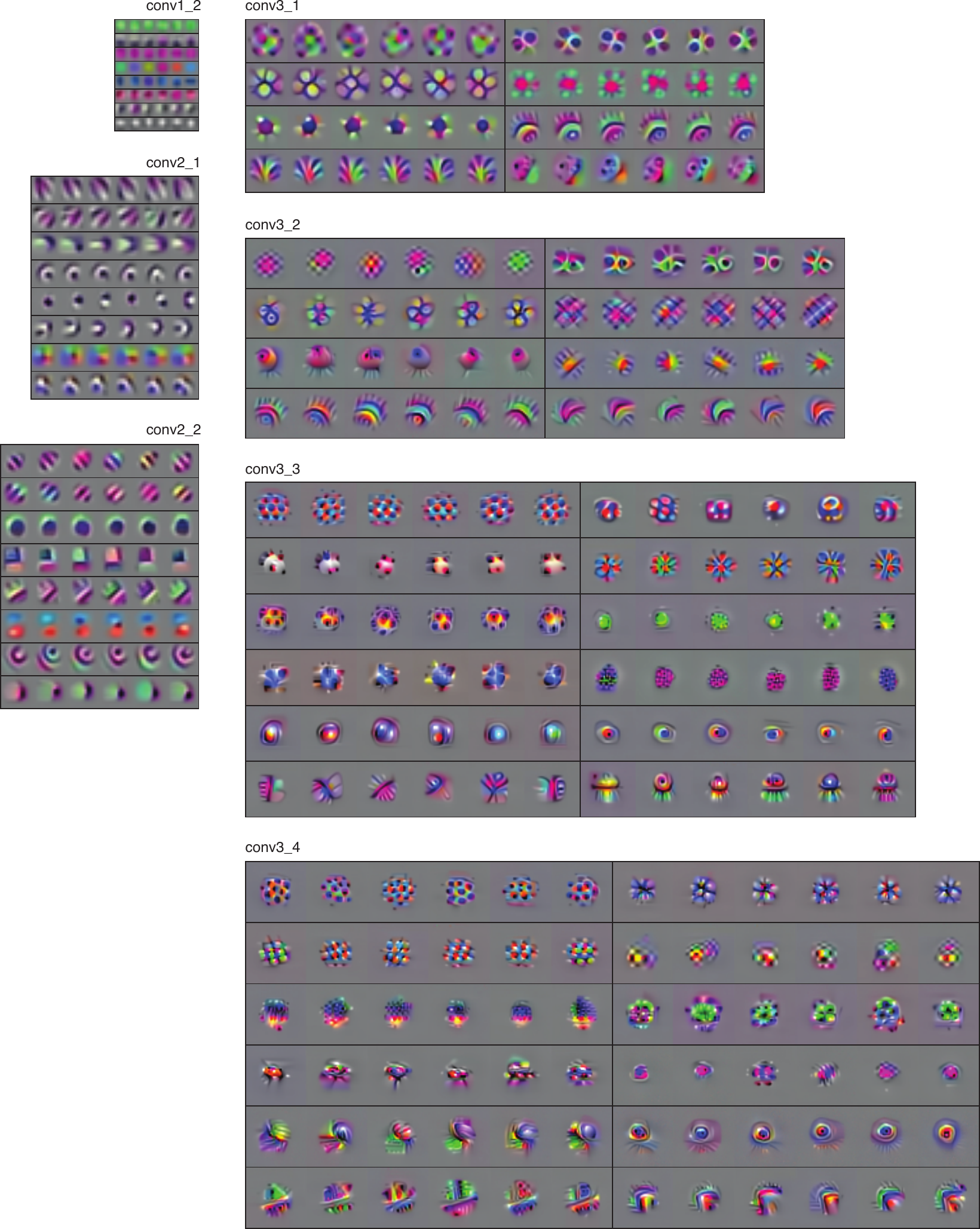}
\caption{Invariant subspaces of a selection of units in convolutional layers conv1\_2 to conv3\_4 of VGG-19. Each horizontal block of six images represents one unit. It contains the six maximally diverse images resulting in an activation of the unit above 90\% of its maximum. 
Images for higher layers are scaled up slightly to improve visibility, but the pixel sizes are not matched across layers (lower layers have comparably larger pixels).
}
\label{fig:vgg_units}
\end{figure*}

The two units shown here are representative of a larger number of units in various layers of VGG-19 (see Fig.~\ref{fig:vgg_units} and Sect. ~2 from Supp. for more examples). As we will quantitatively show below, they lie on two extremes of a spectrum along which we can characterize low- and intermediate-level units.

\subsection{Quantification of phase invariance (textures)}

So far, we have described texture and shape units only qualitatively. We therefore developed metrics to quantify these properties more systematically. We start by quantifying phase invariance, the property that characterizes texture detectors.

\begin{figure}[t!]
\begin{center}
   \includegraphics[width=\linewidth]{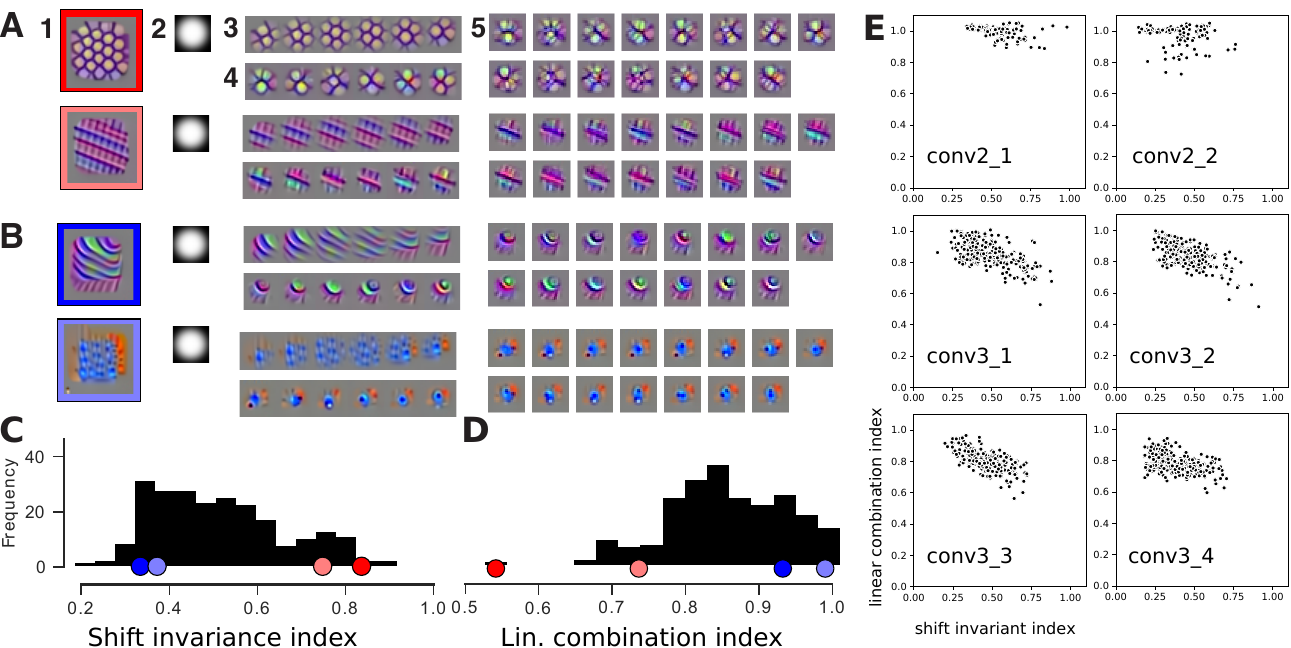}
\end{center}
    \vspace{-8pt}
    \caption{
        Quantification of invariances in VGG19. For layer conv3\_1, examples of texture (\textbf{A}) and shape (\textbf{B}) units.
        \textbf{Left:} Phase invariance. 
        We optimize a texture (\textbf{1}) to maximize the average activation of all windowed (\textbf{2}) crops (\textbf{3}). The mask has the form $\exp\left(-\left(r/\sigma\right)^4\right)$ where $r = \sqrt{x^2 + y^2}$. We picked $\sigma$ so that the ratio between the unit's receptive field and $\sigma$ is $\sim 2.5$. (\textbf{4}): individual images maximizing the unit's activation.
        \textbf{Right:} Invariance to local deformations is supported by features that locally form quadrature pairs. Linear combinations (\textbf{5}) of templates (\textbf{4}) produce images with high activations.
        \textbf{C.} Histogram of the phase invariance (examples from A+B labeled).
        \textbf{D.} Histogram of metric measuring invariance to local deformations.
        \textbf{E.} Scatter plot of the two metrics (shift invariant index and linear combination index) for all units at each convolutional layer of VGG19.
        }
\label{fig:phase}
\end{figure}

While shift \emph{equivariance} is built into CNNs, phase \emph{invariance} of individual units has to our knowledge not been reported. A perfectly phase-invariant unit would maintain a high activation when presented with shifted versions of its preferred texture. Therefore, to quantify phase invariance, we optimize an image twice as large as the unit's receptive field such that the average activation of all possible windowed crops from this image is maximized (Fig.~\ref{fig:phase}A, 1--4). Indeed, for a decent number of units we had qualitatively labeled as `texture detectors,' the crops generated in this way (Fig.~\ref{fig:phase}A,\,3) resemble the templates we synthesized earlier (Fig.~\ref{fig:phase}A,\,4) and elicit similarly high activations (Fig.~\ref{fig:phase}C). On the other hand, `shape-selective' units expect certain structures in specific locations within their receptive field. Generating a texture where arbitrary crops are highly activating is not possible for these units (Fig.~\ref{fig:phase}B).

To quantify this intuitive argument, we defined shift invariance as the ratio between the average activation of all crops from the larger texture and the average activation of the diverse templates produced earlier (see example histogram in Fig.~\ref{fig:phase}C, for conv3\_1). Indeed, the units labeled as phase-invariant (Fig.~\ref{fig:phase}A), maintain a high activations despite arbitrary phase shifts, while the activation of the shape-selective units (Fig.~\ref{fig:phase}B) drops substantially (Fig.~\ref{fig:phase}C).

Note that synthesizing a larger image by maximizing all crops is similar to maximizing an entire channel's activity (i.\,e. feature map) for a sufficiently large input image, an approach other authors have taken for feature visualization \cite{O1}. Although insightful in many occasions, the drawback is that this procedure often occludes shape selectivity. For instance, the first unit in Fig.~\ref{fig:phase}B is selective to a circular pattern in the top-right with rays pointing towards the bottom-left when maximized individually. However, the resulting texture looks like a field of oriented edges, thus missing the crucial pattern that drives this unit.

\subsection{Tolerance to local deformations (shapes)}

The second invariance we identify is tolerance to local deformations. A closer look at some examples (e.\,g. Fig ~\ref{fig:two_examples}, right; Fig.~\ref{fig:phase}B, top) reveals that some of the units have local tolerance for phase changes. The patterns these units are tuned for can be locally built by spatially arranging multiple complex-cell-like quadrature pairs. This would suggest, that -- although mapped into a nonlinear feature space -- linear combinations of the `template' images spanning the invariant subspace should highly activate these units as well. We illustrate this seemingly counter-intuitive hypothesis with a toy example and then show how it applies to CNNs.

\begin{figure}[t!]
\begin{center}
   \includegraphics[width=90mm]{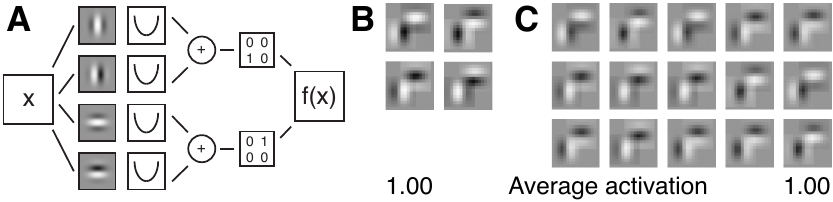}
\end{center}
    \vspace{-8pt}
    \caption{Toy example (\textbf{A}) where linear combinations (\textbf{C}) of highly activating images~(\textbf{B}) are also highly activating. It detects a top-left corner by combining two complex cells.}
\label{fig:toy}
\end{figure}

Consider the following example comprised of two complex cells arranged such that they detect a top-left corner (Fig.~\ref{fig:toy}). The unit allows for individually shifting up or down the horizontal edge, and left or right the vertical edge. Each of the two edges is detected by an energy model of a complex cell (Fig.~\ref{fig:toy}A), each at a defined location within the receptive field. Accordingly, the highly activating template images are made up of combinations of odd and even Gabors (Fig.~\ref{fig:toy}B) and any linear combination of them is again a highly activating image (Fig.~\ref{fig:toy}C).

To quantify whether the same property holds for VGG units, we computed the average activation level of linear combinations of the maximally activating images. Specifically, we took the averages (in pixel space) of all 15~pairs of templates (Fig.~\ref{fig:phase}A.5), renormalized them to the same norm as the templates and compared their average activation to that of the templates. For `texture-selective' units this procedure deteriorates the clear texture patterns revealed by the templates (see for instance Fig.~\ref{fig:phase}A.5). Accordingly, the unit's activation level to these images drops substantially (Fig.~\ref{fig:phase}D, red+orange). We quantify this drop by computing a linear combination index, defined as the ratio between the average activation of average-image pairs and the average activation of the diverse templates. Units tuned to shape patterns that are tolerant to local transformations give average-pairs that are fairly similar to the original templates, producing a high linear combination index.

\subsection{Characterization of invariances across layers}

We have identified two metrics that quantify two different forms of invariance in VGG units. Our examples from Fig.~\ref{fig:phase} suggest that these two types of invariance are anticorrelated. As this does not have to be the case a priori -- a complex cell would score high on both metrics -- we asked whether this was just due to our selection of examples or whether it holds more generally across layers. Indeed, shift invariance and tolerance to local deformations appear to be anticorrelated across a wide range of layers (Fig.~\ref{fig:phase}E; conv3 in particular). We also observe that higher layers tend to be less shift-invariant than lower ones (e.\,g. compare within conv3 in Fig.~\ref{fig:phase}E).


\section{Diverse visualizations of early layers of ResNet-50}

\begin{figure}[t!]
\begin{center}
  \includegraphics[width=\linewidth]{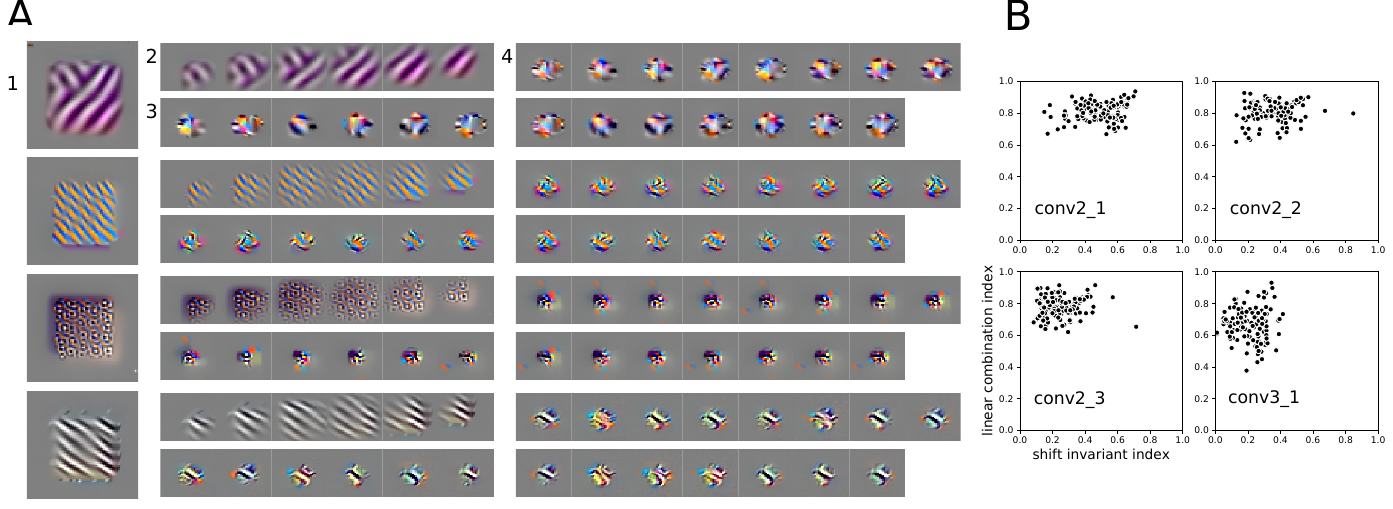}
\end{center}
    \vspace{-8pt}
    \caption{ResNet-50 results. \textbf{A}, Example units of block conv2\_3 (compare to Fig.\ref{fig:phase}). We noticed that maximizing windowed crops (2) of a big texture (1) are largely different from the maximizing templates (3). 15 template-pair averages (4) are on the other hand highly activating and similar to the templates. \textbf{B}, Scatter plot of the two metrics proposed for said layers of ResNet-50.}
\label{fig:resnet}
\end{figure}

To test whether our results so far are properties of VGG-19 or apply more generally to CNNs trained on ImageNet, we also applied our methods to ResNet-50~\cite{he2016deep}. We considered its early layers up to conv3\_1 (fourth block), which have receptive field sizes comparable to the layers we studied in VGG-19. We first synthesized diverse image batches with different diversity penalties and found a similar trade-off between activation and diversity as found before (see Sect.~3 in Suppl. Material). However, for the $\lambda$ that evoked at least 90\% of the maximal responses we observed on average a smaller diversity compared to that of VGG-19 units. We then ran our analysis to identify both phase and shape invariance and surprisingly found a much reduced number of phase-invariant units compared to VGG-19 (Fig.~\ref{fig:resnet}): there are basically no ResNet-50 units for which the crops from the optimal texture look like the optimized templates (e.g. Fig.~\ref{fig:resnet}A,2+3). On the other hand, template-pair averages do not appear to qualitatively deviate from the synthesized templates (Fig.~\ref{fig:resnet}A,4) indicating a strong presence of tolerance to local changes. The two metrics introduced above confirm this observation quantitatively: the distribution of shift invariance indices is shifted towards zero in ResNet-50 layers (Fig.~\ref{fig:resnet}B) with respect to those in VGG-19.

This is a very interesting finding, because it shows that the different architectures learn quite different features in their early layers despite both being trained on ImageNet and achieving comparable classification accuracy. Thus, our novel approach to feature visualization helped us identify strong representational differences in the canonical directions between two architectures that would not have been observed with conventional activity maximization

\section{Phase invariance in Primary visual cortex (V1)}

\begin{figure}[t!]
\begin{center}
  \includegraphics[width=\linewidth]{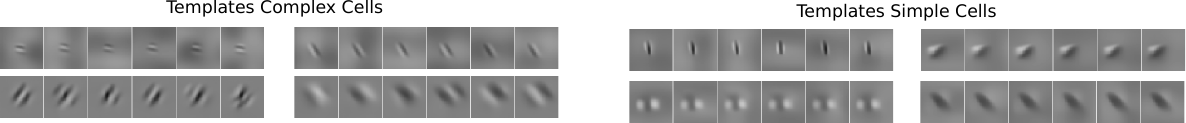}
\end{center}
    \vspace{-8pt}
    \caption{Subspaces of V1 cells. Complex (left) and simple (right) cells}
\label{fig:v1}
\end{figure}

As a final practical use case, we applied our method to a three-layer CNN that has been trained to predict neural responses in V1 when monkeys are shown natural images (data from \cite{C1}; see also their Fig.~3). Our method unveils the known cell types -- simple: phase-selective and complex: phase-invariant (Fig \ref{fig:v1}). Although complex cells can also be identified using specifically designed stimuli or analysis methods relying on quadratic features (e.\,g. spike-triggered covariance \cite{rust2005spatiotemporal}), our non-parametric approach could in principle also uncover other types of invariance that are not captured by quadratic features. Given that we see no such additional invariances, there are likely no other major features V1 cells are invariant to -- a conclusion that could not be drawn using parametric approaches.

\section{Conclusion}

Motivated by early vision in the brain, we investigated the response invariances in the early to intermediate convolutional layers of DNNs.
We found that units in early layers of VGG-19 show invariance to global texture-preserving transformations and invariance to local shape-preserving transformations. In contrast, ResNet-50 does not exhibit the same degree of shift invariance. This difference could explain why practitioners working on texture synthesis and style transfer observe that the features of VGG work substantially better than those of more modern architectures such as residual networks.


We conclude that these methods not only give new insights into the computations performed by DNNs and how they compare with other architectures, but also constitutes an important step towards a unified language for describing neural representations in both biological and computer vision.

\subsubsection*{Acknowledgements.} We thank Jonas Rauber and Andreas Tolias for useful discussions. This work was supported by the German Research Foundation (DFG) grant EC 479/1-1 to A.S.E. The International Max Planck Research School for Intelligent Systems (IMPRS-IS) supported S.A.C. The work was also supported by IARPA via Department of Interior (DoI) contract D16PC00003.

\bibliographystyle{splncs04}
\bibliography{references.bib}

\begin{thebibliography}{10}
\providecommand{\url}[1]{\texttt{#1}}
\providecommand{\urlprefix}{URL }
\providecommand{\doi}[1]{https://doi.org/#1}

\bibitem{A1}
Adelson, E.H., Bergen, J.R.: Spatiotemporal energy models for the perception of
  motion. J. Opt. Soc. Am. A  \textbf{2}(2),  284--299 (1985).
  \doi{10.1364/JOSAA.2.000284}

\bibitem{berkes2005slow}
Berkes, P., Wiskott, L.: Slow feature analysis yields a rich repertoire of
  complex cell properties. Journal of vision  \textbf{5}(6), ~9--9 (2005)

\bibitem{bethge2007unsupervised}
Bethge, M., Gerwinn, S., Macke, J.H.: Unsupervised learning of a steerable
  basis for invariant image representations. In: Human Vision and Electronic
  Imaging XII. vol.~6492, p. 64920C. International Society for Optics and
  Photonics (2007)

\bibitem{C1}
Cadena, S.A., Denfield, G.H., Walker, E.Y., Gatys, L.A., Tolias, A.S., Bethge,
  M., Ecker, A.S.: Deep convolutional models improve predictions of macaque v1
  responses to natural images. bioRxiv  (2017). \doi{10.1101/201764}

\bibitem{Ca1}
Cadieu, C.F., Hong, H., Yamins, D.L., Pinto, N., Ardila, D., Solomon, E.A.,
  Majaj, N.J., DiCarlo, J.J.: Deep neural networks rival the representation of
  primate {IT} cortex for core visual object recognition. PLoS computational
  biology  \textbf{10}(12),  e1003963 (2014), 00152

\bibitem{E1}
Erhan, D., Bengio, Y., Courville, A., Vincent, P.: Visualizing higher-layer
  features of a deep network. Tech. Rep.~1341, University of Montreal (Jun
  2009), also presented at the ICML 2009 Workshop on Learning Feature
  Hierarchies, Montr{\'{e}}al, Canada.

\bibitem{Ga2}
Gatys, L., Ecker, A.S., Bethge, M.: Texture synthesis using convolutional
  neural networks. In: Advances in Neural Information Processing Systems. pp.
  262--270 (2015)

\bibitem{Ga1}
Gatys, L.A., Ecker, A.S., Bethge, M.: Image style transfer using convolutional
  neural networks. In: Proceedings of the IEEE Conference on Computer Vision
  and Pattern Recognition. pp. 2414--2423 (2016)

\bibitem{G1}
Goodfellow, I., Lee, H., Le, Q.V., Saxe, A., Ng, A.Y.: Measuring invariances in
  deep networks. In: Advances in neural information processing systems. pp.
  646--654 (2009)

\bibitem{Gu1}
G{\"u}{\c c}l{\"u}, U., van Gerven, M.A.J.: Deep neural networks reveal a
  gradient in the complexity of neural representations across the ventral
  stream. Journal of Neuroscience  \textbf{35}(27),  10005--10014 (2015).
  \doi{10.1523/JNEUROSCI.5023-14.2015}

\bibitem{he2016deep}
He, K., Zhang, X., Ren, S., Sun, J.: Deep residual learning for image
  recognition. In: Proceedings of the IEEE conference on computer vision and
  pattern recognition. pp. 770--778 (2016)

\bibitem{Hu1}
Hubel, D.H., Wiesel, T.N.: Receptive fields, binocular interaction and
  functional architecture in the cat's visual cortex. The Journal of physiology
   \textbf{160}(1), ~106 (1962), 09139

\bibitem{Ki1}
Kingma, D.P., Ba, J.: Adam: A method for stochastic optimization. arXiv
  preprint arXiv:1412.6980  (2014)

\bibitem{Kr1}
Kriegeskorte, N.: Deep neural networks: A new framework for modeling biological
  vision and brain information processing. Annual Review of Vision Science
  \textbf{1}(1),  417--446 (2015). \doi{10.1146/annurev-vision-082114-035447}

\bibitem{lies2014slowness}
Lies, J.P., H{\"a}fner, R.M., Bethge, M.: Slowness and sparseness have
  diverging effects on complex cell learning. PLoS computational biology
  \textbf{10}(3),  e1003468 (2014)

\bibitem{M2}
Mahendran, A., Vedaldi, A.: Understanding deep image representations by
  inverting them. In: Proceedings of the IEEE conference on computer vision and
  pattern recognition. pp. 5188--5196 (2015)

\bibitem{M1}
Mahendran, A., Vedaldi, A.: Visualizing deep convolutional neural networks
  using natural pre-images. International Journal of Computer Vision
  \textbf{120}(3),  233--255 (2016)

\bibitem{N1}
Nguyen, A., Clune, J., Bengio, Y., Dosovitskiy, A., Yosinski, J.: Plug \& play
  generative networks: Conditional iterative generation of images in latent
  space. In: CVPR. vol.~2, p.~7 (2017)

\bibitem{N4}
Nguyen, A., Dosovitskiy, A., Yosinski, J., Brox, T., Clune, J.: Synthesizing
  the preferred inputs for neurons in neural networks via deep generator
  networks. In: Advances in Neural Information Processing Systems. pp.
  3387--3395 (2016)

\bibitem{N3}
Nguyen, A., Yosinski, J., Clune, J.: Deep neural networks are easily fooled:
  High confidence predictions for unrecognizable images. In: The IEEE
  Conference on Computer Vision and Pattern Recognition (June 2015)

\bibitem{N2}
Nguyen, A.M., Yosinski, J., Clune, J.: Multifaceted feature visualization:
  Uncovering the different types of features learned by each neuron in deep
  neural networks. Visualization for Deep Learning workshop, ICML  (2016)

\bibitem{O1}
Olah, C., Mordvintsev, A., Schubert, L.: Feature visualization. Distill
  (2017). \doi{10.23915/distill.00007}

\bibitem{van2016conditional}
van~den Oord, A., Kalchbrenner, N., Espeholt, L., Vinyals, O., Graves, A.,
  et~al.: Conditional image generation with pixelcnn decoders. In: Advances in
  Neural Information Processing Systems. pp. 4790--4798 (2016)

\bibitem{oord2016pixel}
Oord, A.v.d., Kalchbrenner, N., Kavukcuoglu, K.: Pixel recurrent neural
  networks. arXiv preprint arXiv:1601.06759  (2016)

\bibitem{R1}
Russakovsky, O., Deng, J., Su, H., Krause, J., Satheesh, S., Ma, S., Huang, Z.,
  Karpathy, A., Khosla, A., Bernstein, M., Berg, A.C., Fei-Fei, L.: {ImageNet
  Large Scale Visual Recognition Challenge}. International Journal of Computer
  Vision (IJCV)  \textbf{115}(3),  211--252 (2015).
  \doi{10.1007/s11263-015-0816-y}

\bibitem{rust2005spatiotemporal}
Rust, N.C., Schwartz, O., Movshon, J.A., Simoncelli, E.P.: Spatiotemporal
  elements of macaque v1 receptive fields. Neuron  \textbf{46}(6),  945--956
  (2005)

\bibitem{Sa1}
Salimans, T., Karpathy, A., Chen, X., Kingma, D.P., Bulatov, Y.: Pixelcnn++: A
  pixelcnn implementation with discretized logistic mixture likelihood and
  other modifications. In: Submitted to ICLR 2017 (2016)

\bibitem{S2}
Simonyan, K., Zisserman, A.: Very deep convolutional networks for large-scale
  image recognition. arXiv preprint arXiv:1409.1556  (2014),
  \url{http://arxiv.org/abs/1409.1556}

\bibitem{Sz1}
Szegedy, C., Zaremba, W., Sutskever, I., Bruna, J., Erhan, D., Goodfellow, I.,
  Fergus, R.: Intriguing properties of neural networks. arXiv preprint
  arXiv:1312.6199  (2013)

\bibitem{theis2015generative}
Theis, L., Bethge, M.: Generative image modeling using spatial lstms. In:
  Advances in Neural Information Processing Systems. pp. 1927--1935 (2015)

\bibitem{theis2012mixtures}
Theis, L., Hosseini, R., Bethge, M.: Mixtures of conditional gaussian scale
  mixtures applied to multiscale image representations. PloS one
  \textbf{7}(7),  e39857 (2012)

\bibitem{W1}
Wei, D., Zhou, B., Torrabla, A., Freeman, W.: Understanding intra-class
  knowledge inside cnn. arXiv preprint arXiv:1507.02379  (2015)

\bibitem{Z1}
Zeiler, M.D., Fergus, R.: Visualizing and understanding convolutional networks.
  In: European conference on computer vision. pp. 818--833. Springer (2014)

\end{thebibliography}

\newpage

\section{Supplementary} 

\subsection{Diversity/activation maximization trade-off VGG19}

As in Figures 4 and 5, we show here the trade-off between diversity and activity maximization for all layers including the natural image prior. Diversity is measured as the minimum $L_2$ distance in feature space between all pairs of synthesized templates. Each curve represents a unit (feature map) of the corresponding layer. The curves connect the average of three optimization runs for a choice of $\lambda$ from Equation 1. The penalty for the natural image prior $\alpha$ was set to $0.0005$ after visual inspection. The curves were normalized to the maximum sum of activations (relative average activation level). On the left: The trade-off between minimum distance and relative average activation. On the right: The same curves normalized to have a unit minimum distance. This facilitated comparison with the network with random weights (black). Here, we show in black a sample of units from a random network with the same architecture as VGG-19. Note that the VGG units exhibit more invariance at each layer than expected from random weights for all studied layers.
\vspace{-1em}
\subsubsection{conv1\_2}.
\vspace{-1em}
\begin{figure}[H]
\centering
\includegraphics[width=80mm]{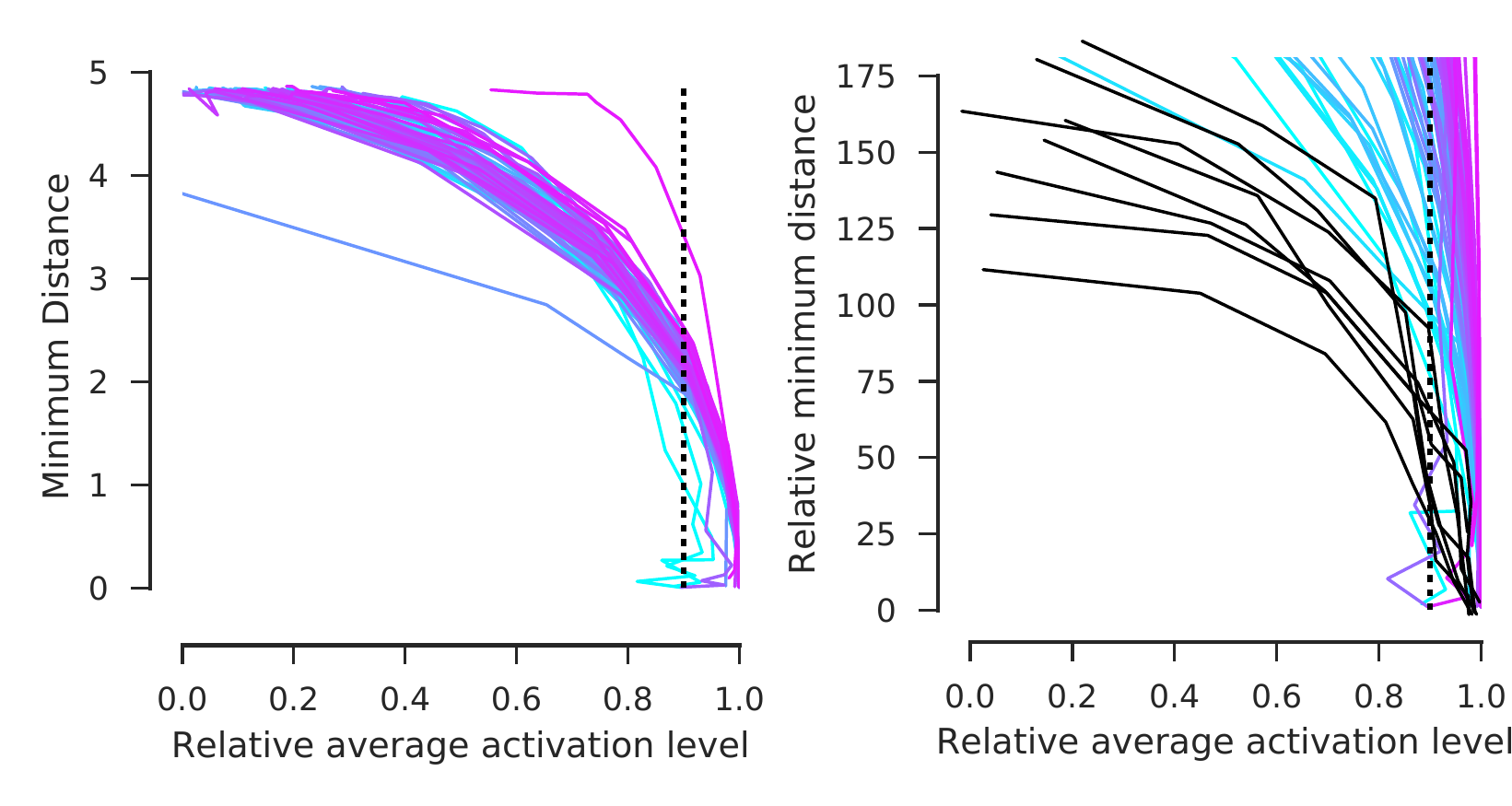}
\end{figure}
\vspace{-1em}

\subsubsection*{conv2\_1}.
\vspace{-1em}
\begin{figure}[H]
\centering
\includegraphics[width=80mm]{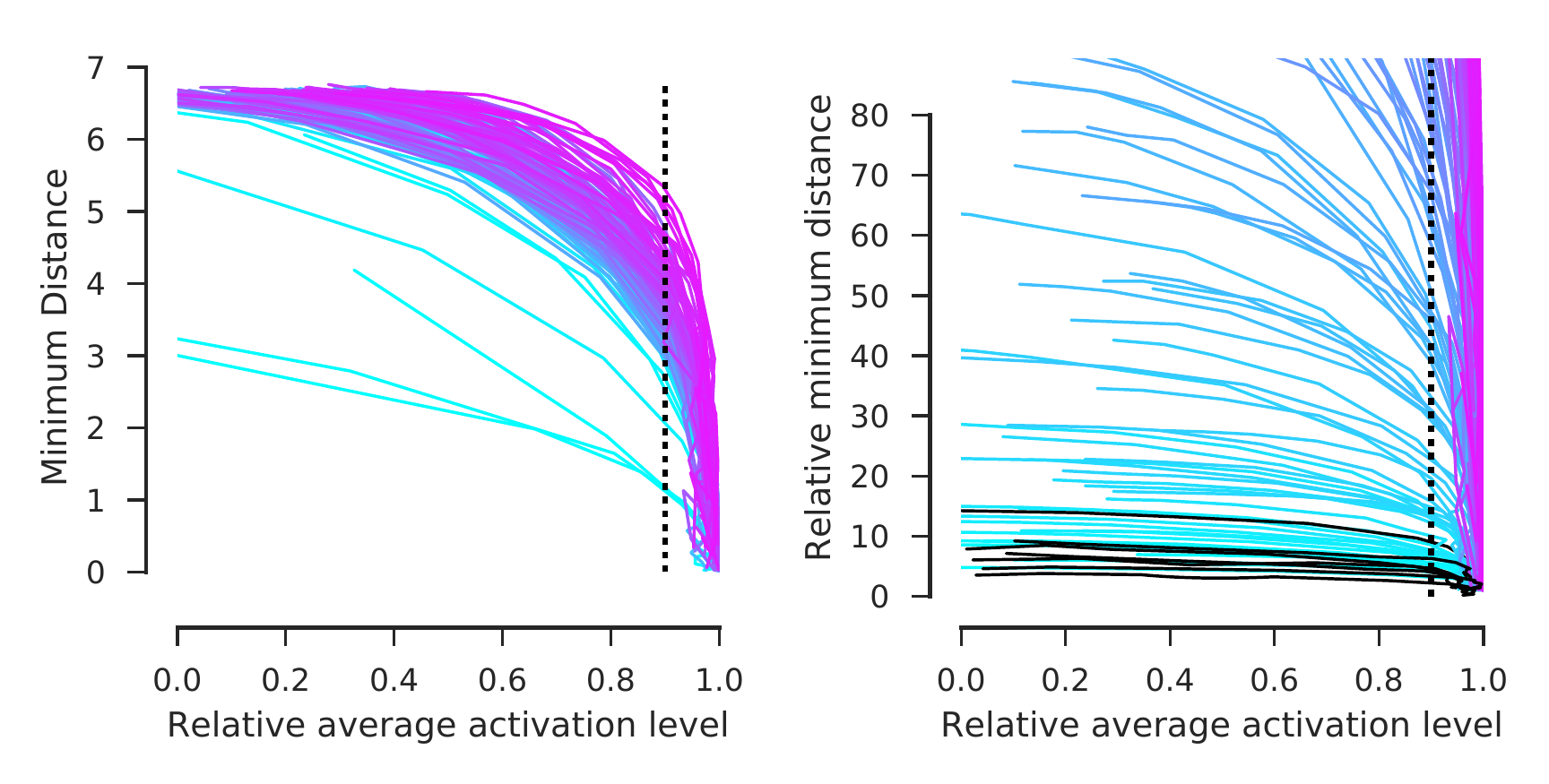}
\end{figure}

\subsubsection*{conv2\_2}.
\begin{figure}[H]
\centering
\includegraphics[width=83mm]{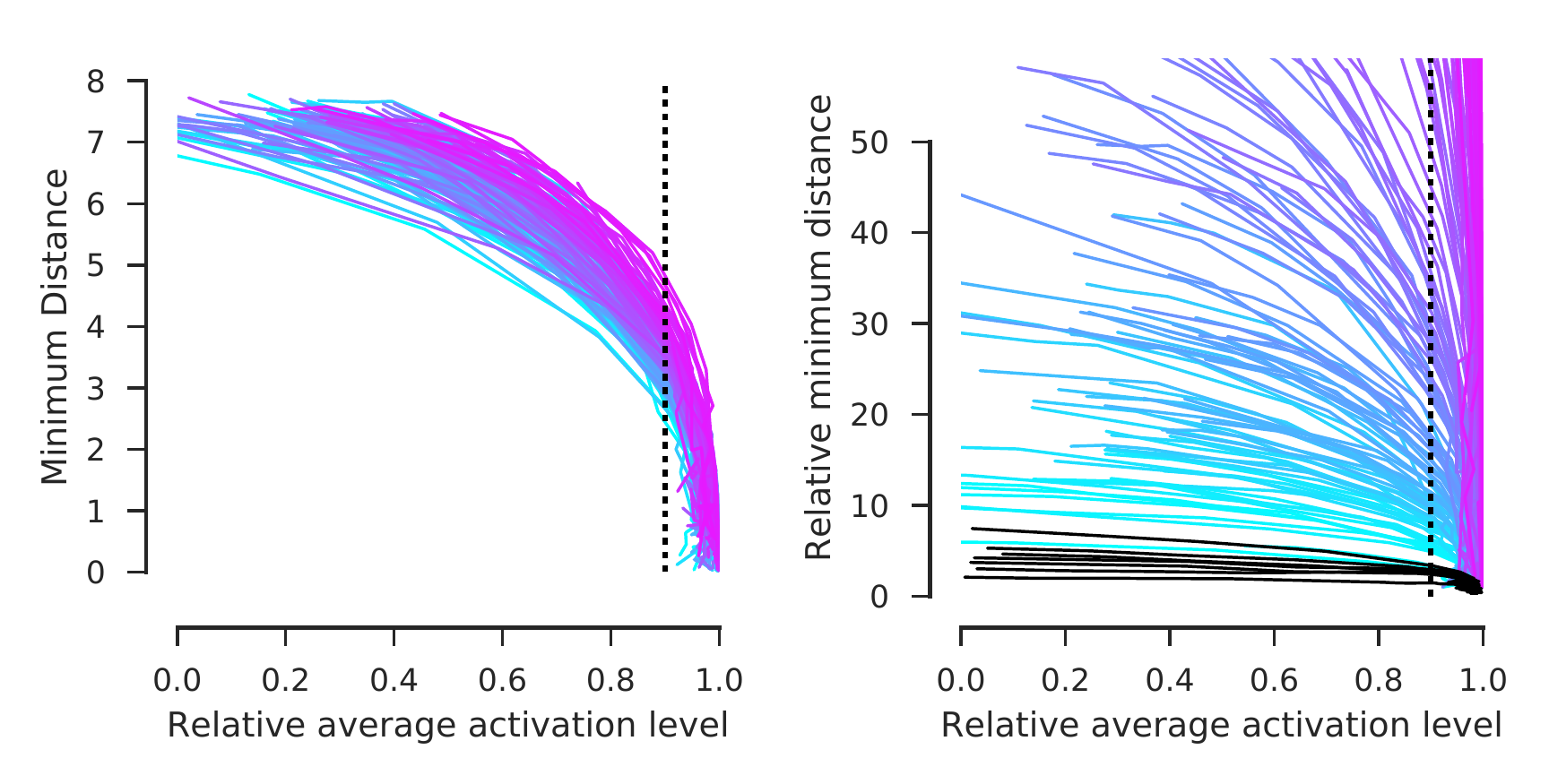}
\end{figure}

\subsubsection*{conv3\_1}.
\begin{figure}[H]
\centering
\includegraphics[width=83mm]{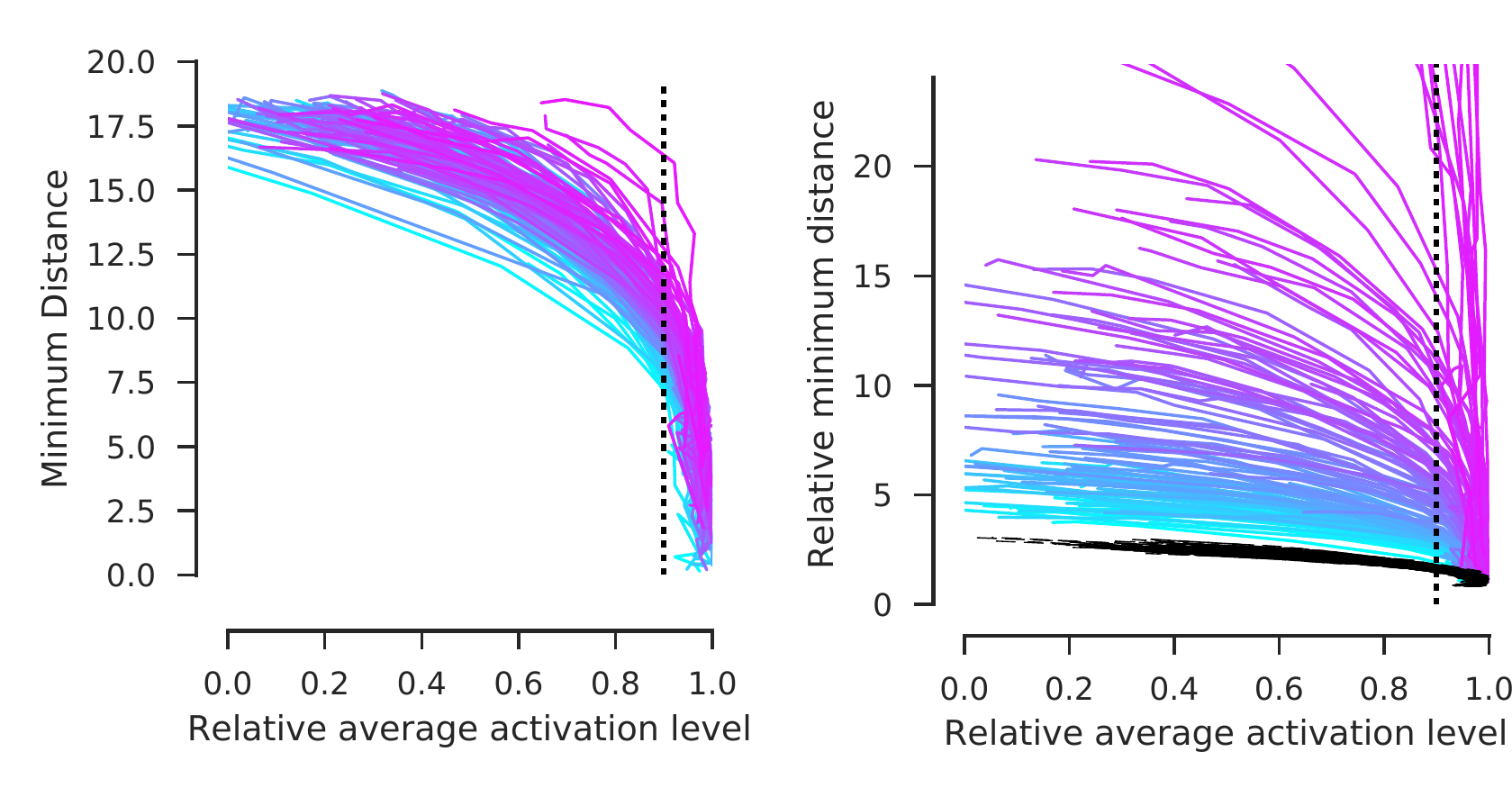}
\end{figure}

\subsubsection*{conv3\_2}.
\begin{figure}[H]
\centering
\includegraphics[width=83mm]{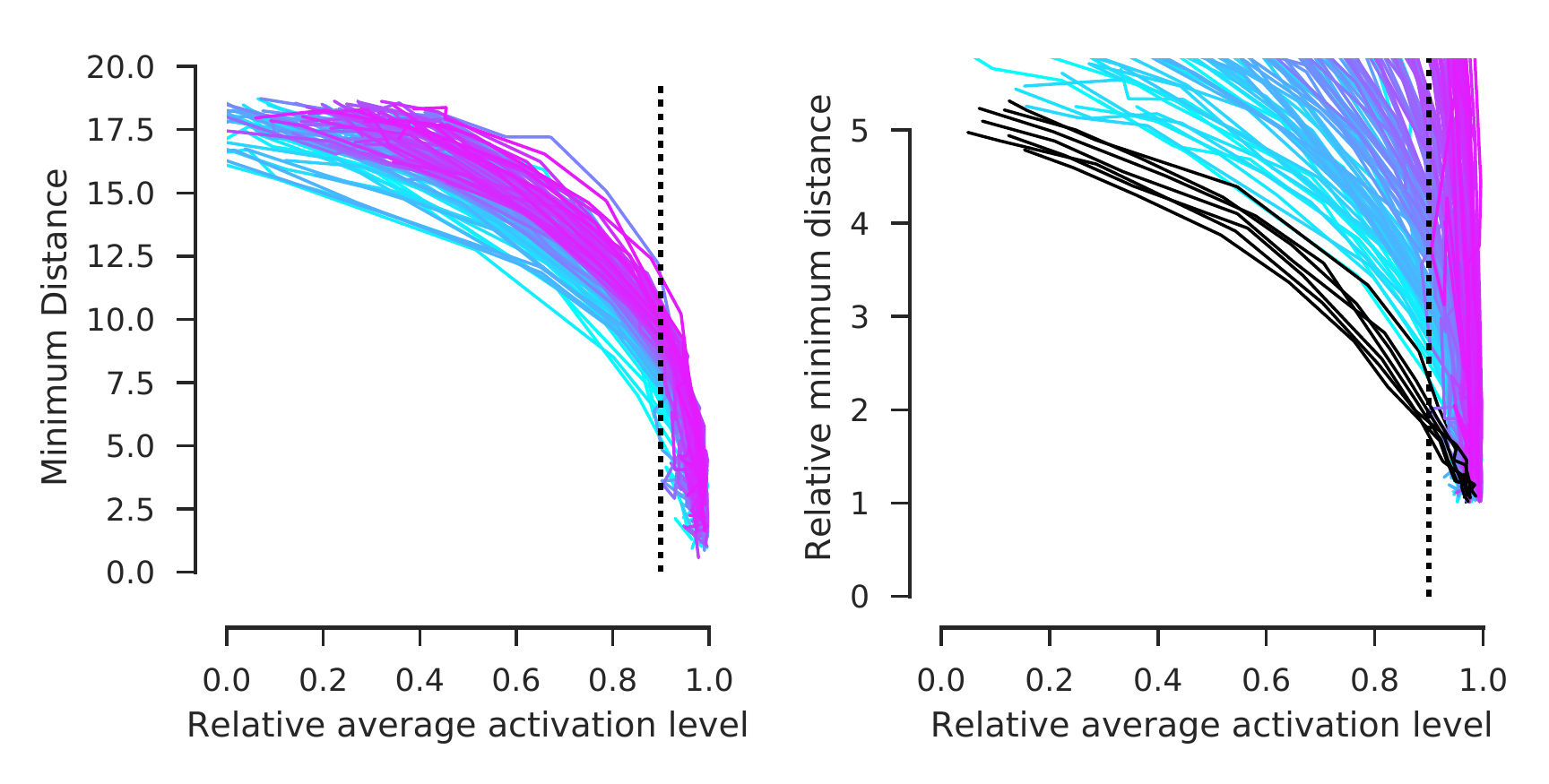}
\end{figure}

\newpage
\subsubsection*{conv3\_3}.
\begin{figure}[H]
\centering
\includegraphics[width=83mm]{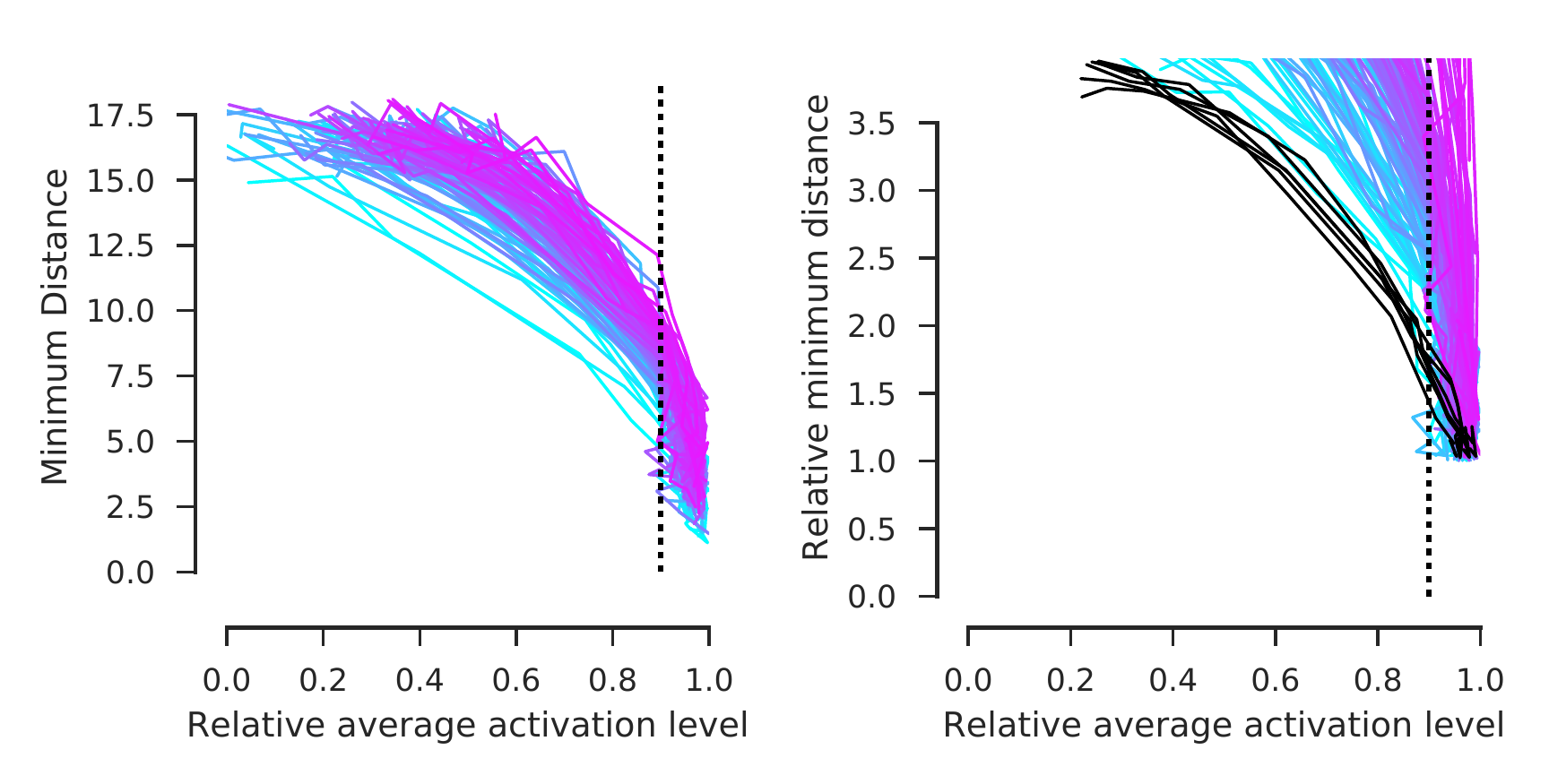}
\end{figure}

\subsubsection*{conv3\_4}.
\begin{figure}[H]
\centering
\includegraphics[width=83mm]{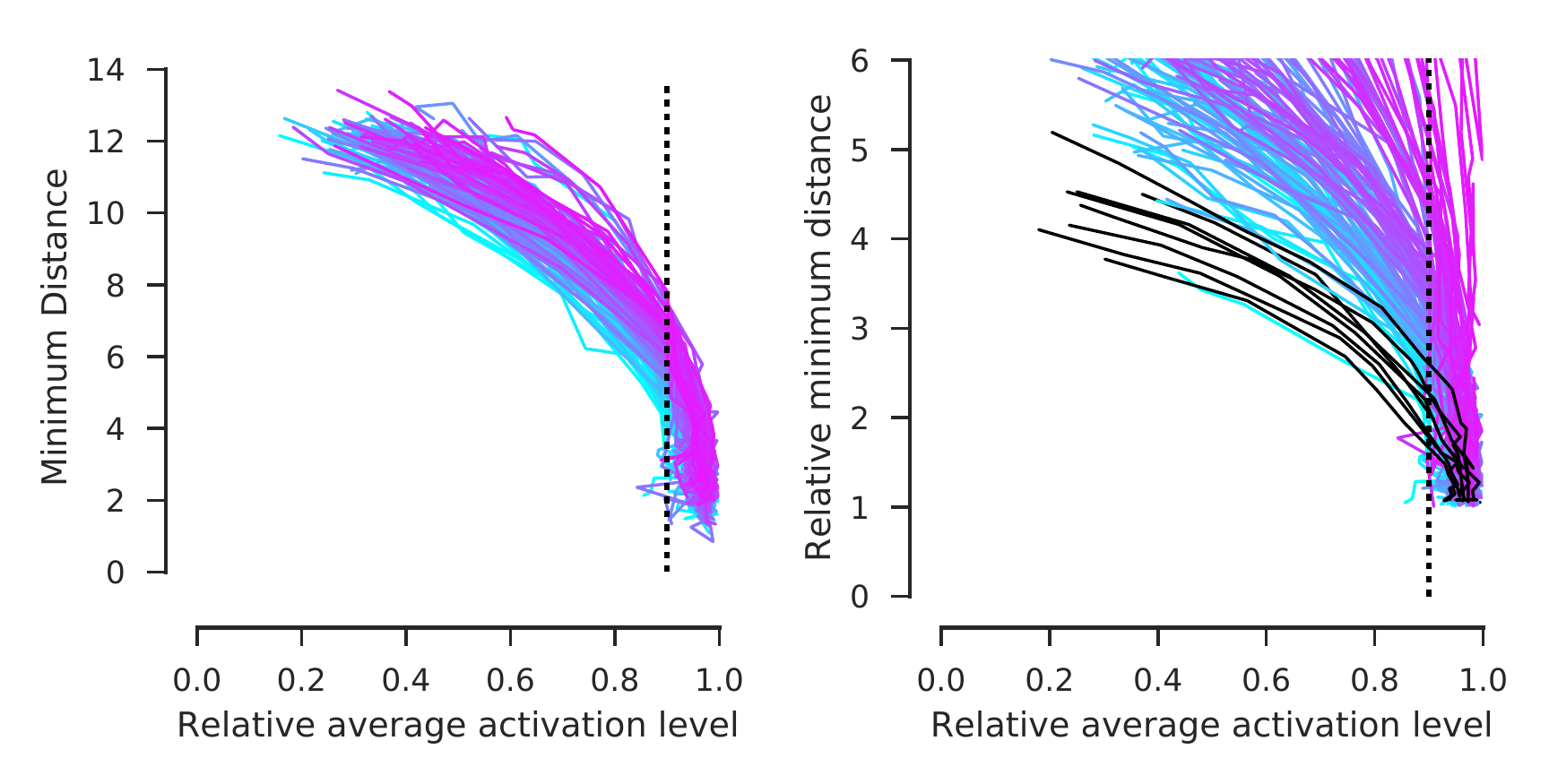}
\end{figure}
\newpage

\subsection{Example invariant subspaces at optimal $\lambda$ for early convolutional layers of VGG-19 }

\subsubsection*{conv1\_2}.
\begin{figure}[H]
\centering
\includegraphics[width=\textwidth]{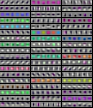}
\end{figure}

\newpage
\subsubsection*{conv2\_1}.
\begin{figure}[H]
\centering
\includegraphics[width=\textwidth]{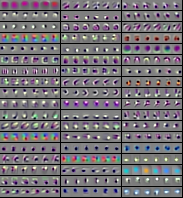}
\end{figure}

\newpage
\subsubsection*{conv2\_2}.
\begin{figure}[H]
\centering
\includegraphics[width=\textwidth]{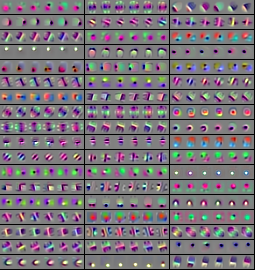}
\end{figure}

\newpage
\subsubsection*{conv3\_1}.
\begin{figure}[H]
\centering
\includegraphics[width=\textwidth]{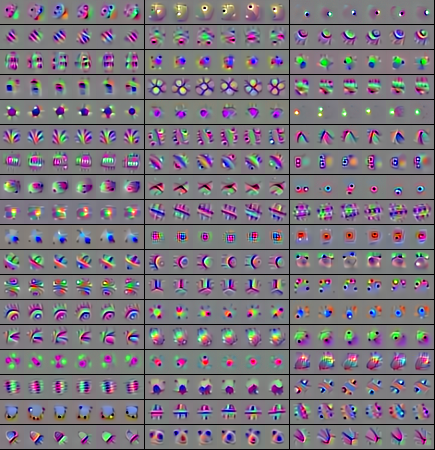}
\end{figure}

\newpage
\subsection*{conv3\_2}.
\begin{figure}[H]
\centering
\includegraphics[width=\textwidth]{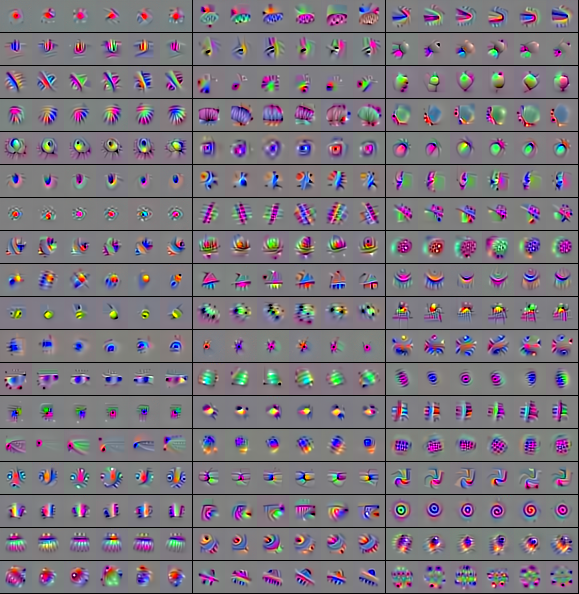}
\end{figure}

\newpage
\subsubsection*{conv3\_3}.
\begin{figure}[H]
\centering
\includegraphics[width=\textwidth]{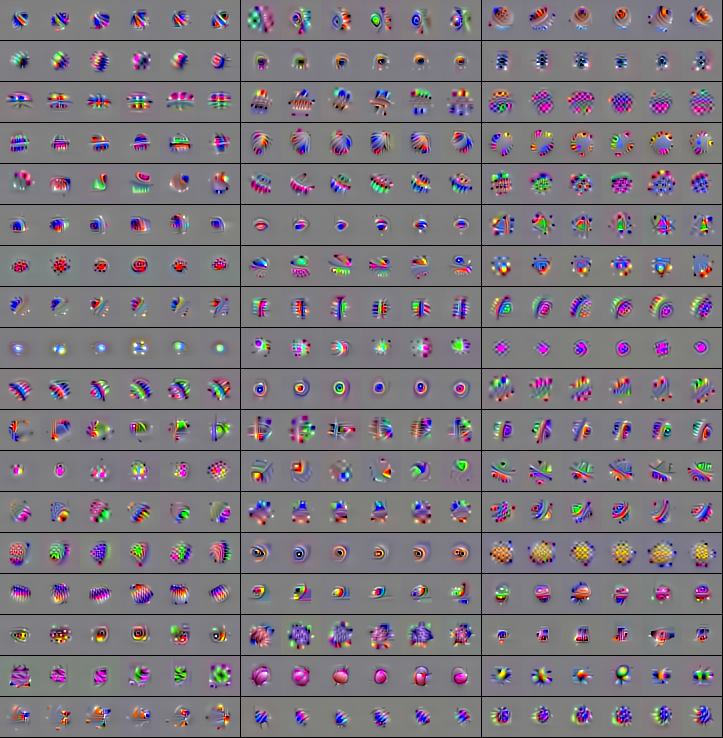}
\end{figure}

\newpage
\subsubsection*{conv3\_4}.
\begin{figure}[H]
\centering
\includegraphics[width=\textwidth]{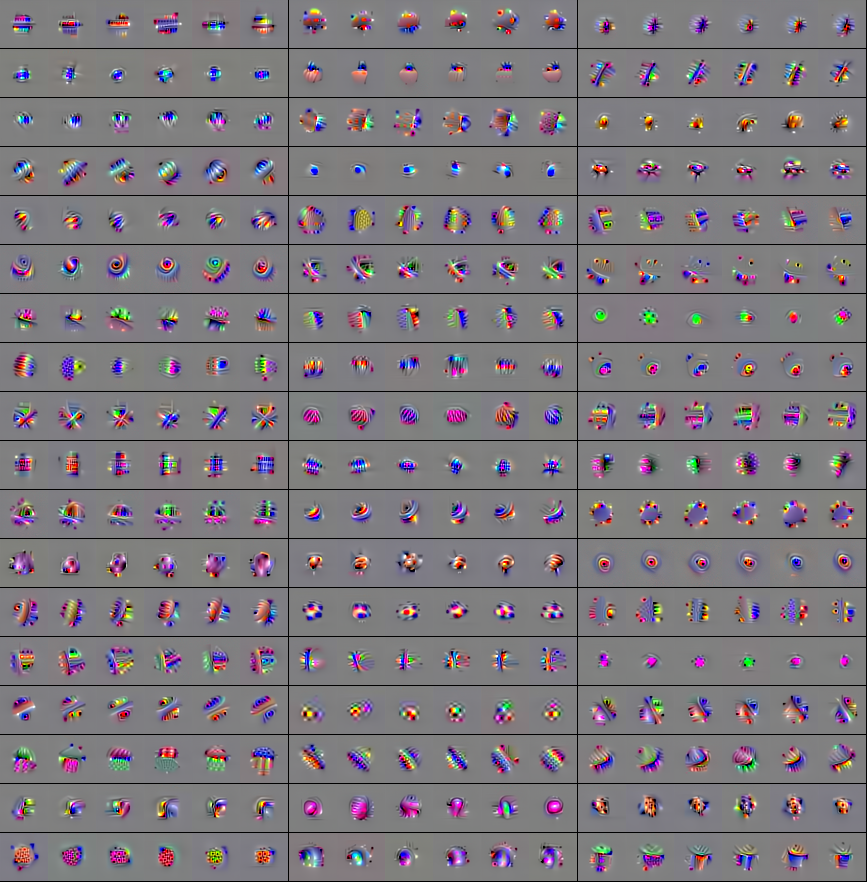}
\end{figure}

\newpage
\subsection{Diversity/activation maximization trade-off ResNet-50}

\subsubsection*{conv2\_1}.
\begin{figure}[H]
\centering
\includegraphics[width=83mm]{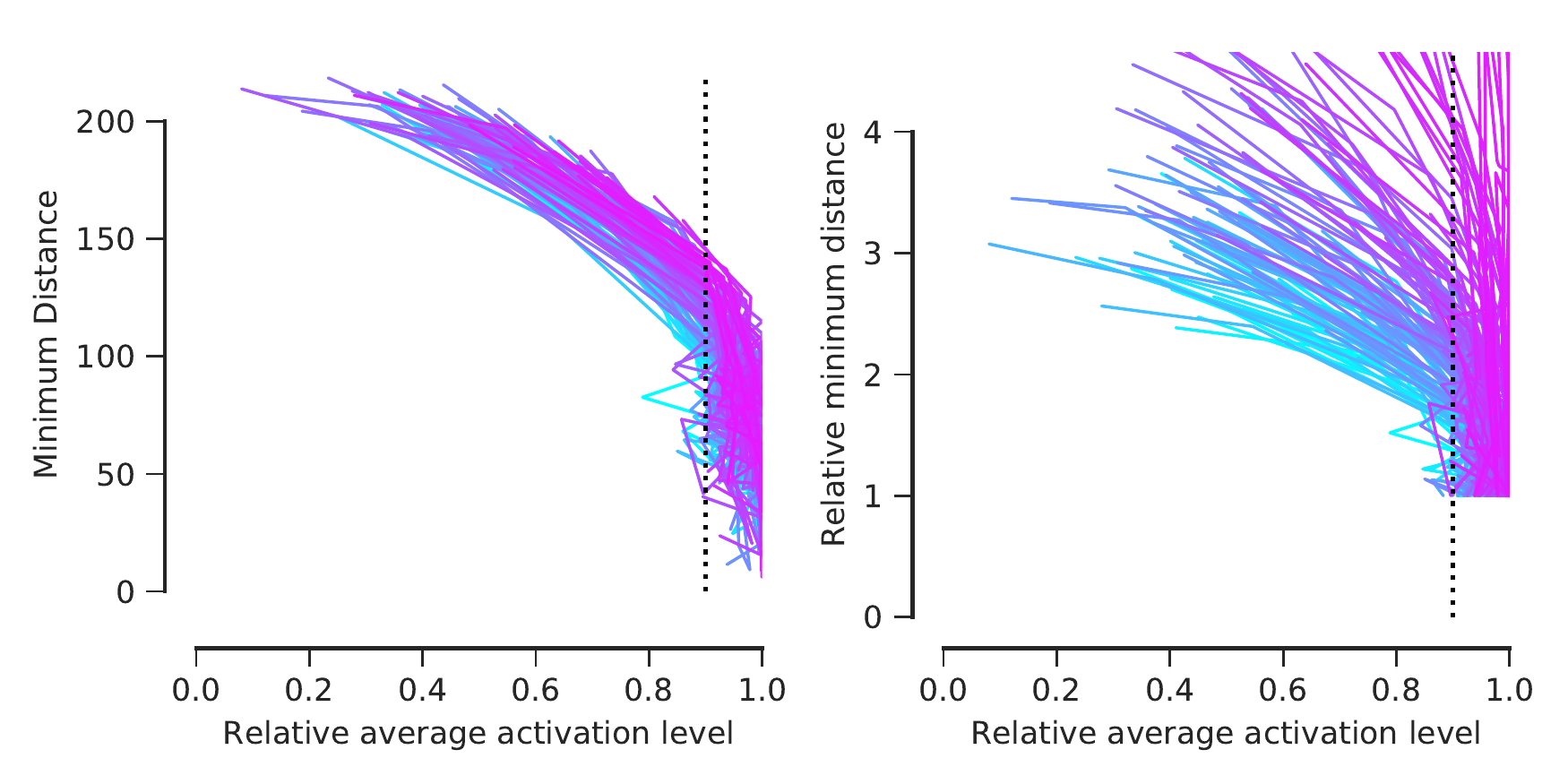}
\end{figure}

\subsubsection*{conv2\_2}.
\begin{figure}[H]
\centering
\includegraphics[width=83mm]{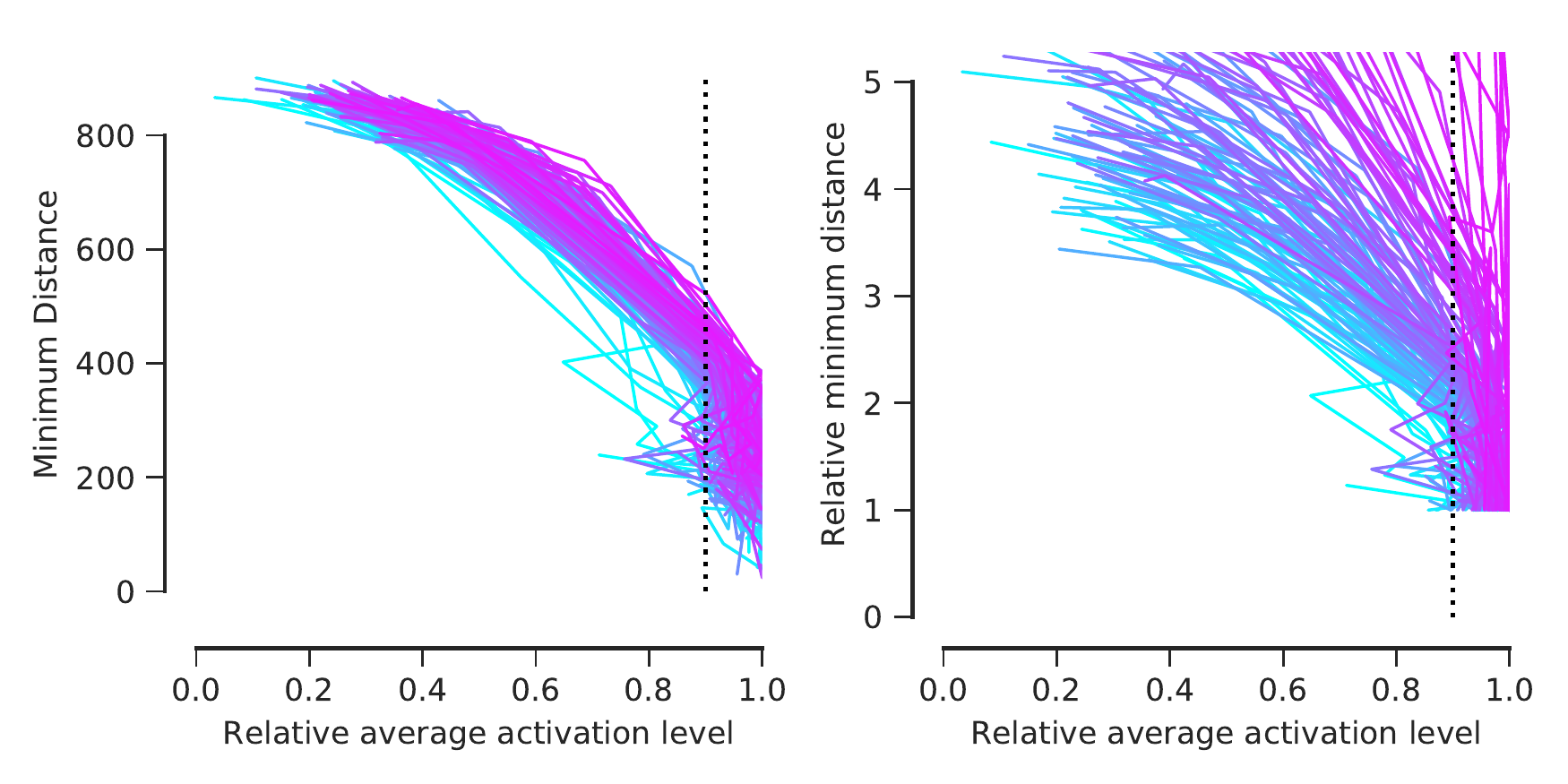}
\end{figure}

\subsubsection*{conv2\_3}.
\begin{figure}[H]
\centering
\includegraphics[width=83mm]{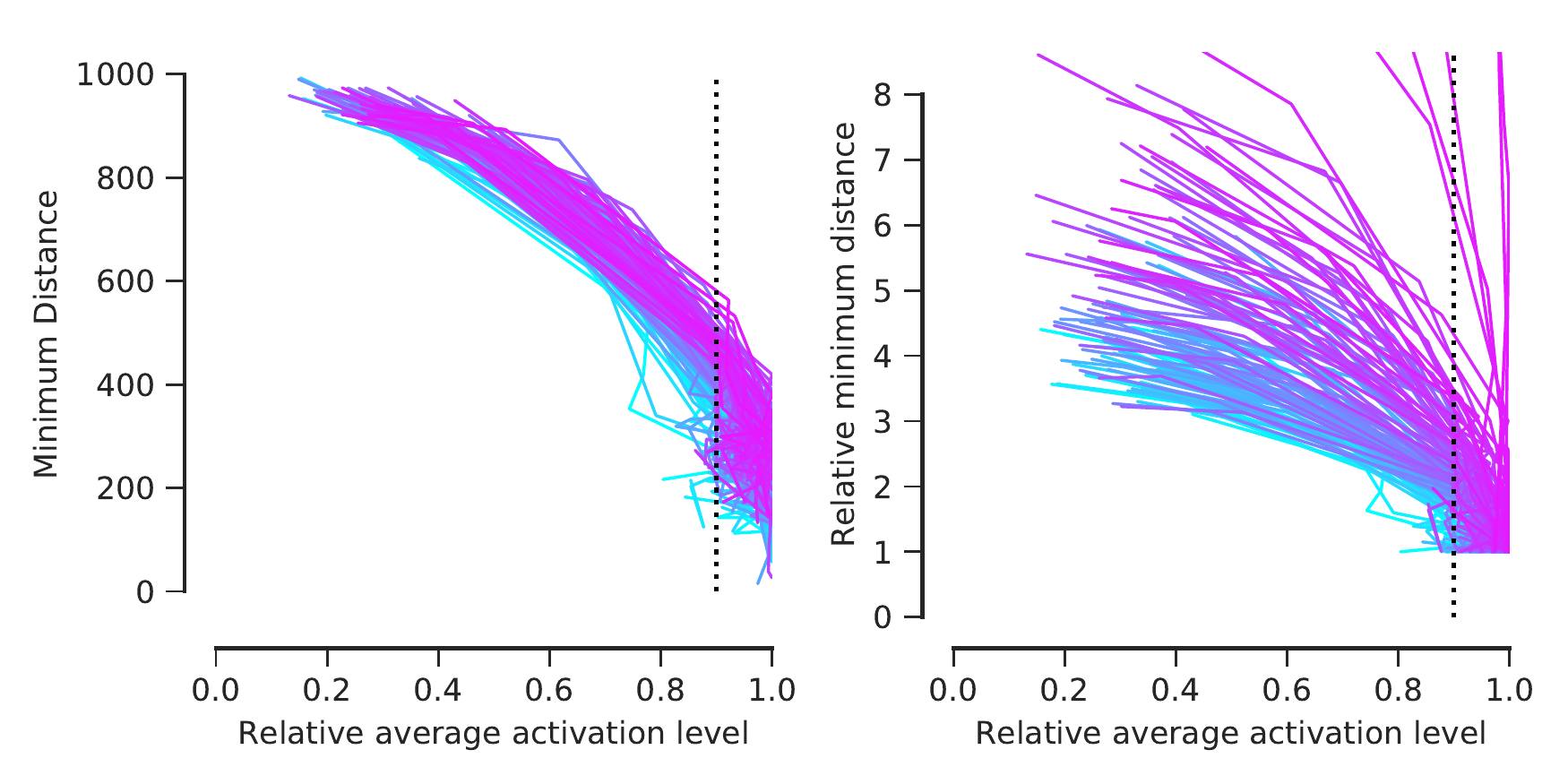}
\end{figure}

\subsubsection*{conv3\_1}.
\begin{figure}[H]
\centering
\includegraphics[width=83mm]{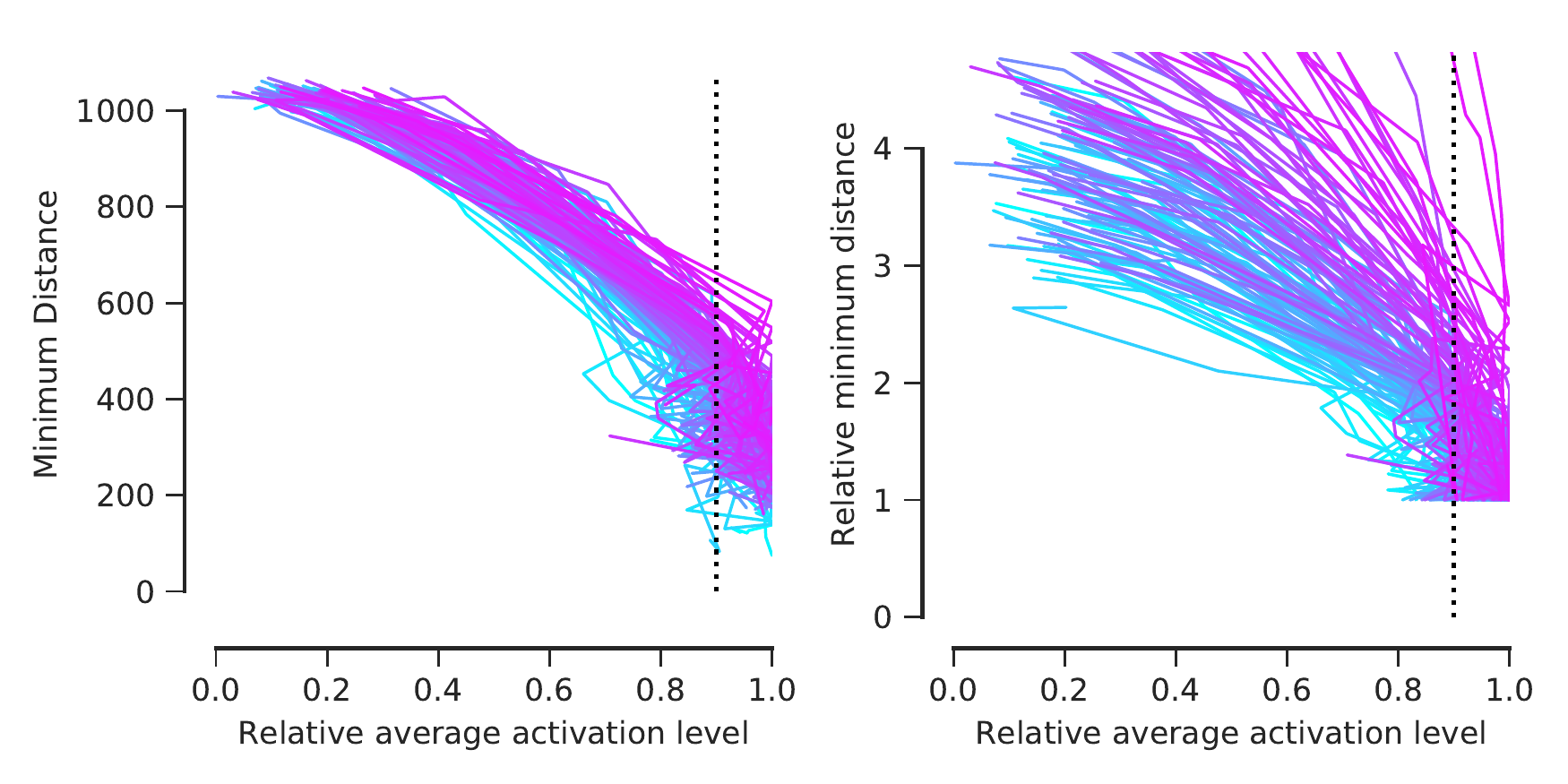}
\end{figure}
\newpage

\subsection{Example invariant subspaces at optimal $\lambda$ for early convolutional layers of ResNet-50}

\subsubsection*{conv2\_1}.
\begin{figure}[H]
\centering
\includegraphics[width=\textwidth]{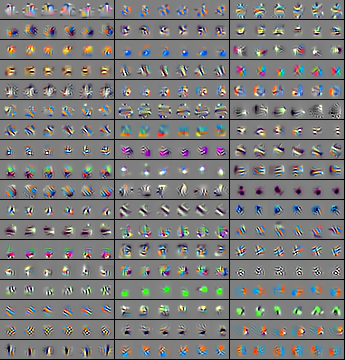}
\end{figure}

\newpage
\subsubsection*{conv2\_2}.
\begin{figure}[H]
\centering
\includegraphics[width=\textwidth]{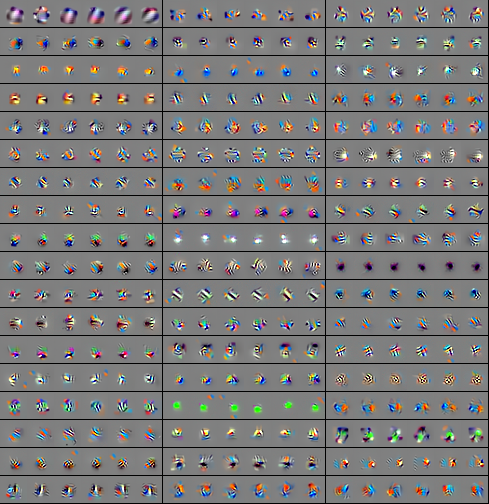}
\end{figure}

\newpage
\subsubsection*{conv2\_3}.
\begin{figure}[H]
\centering
\includegraphics[width=\textwidth]{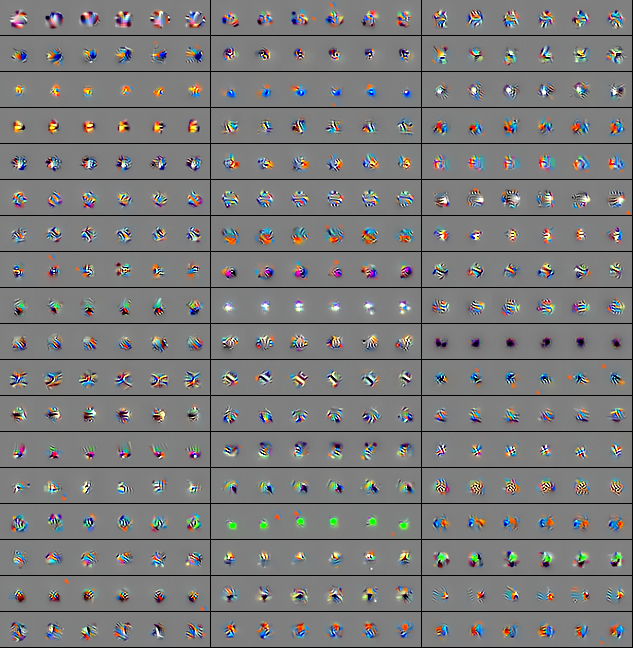}
\end{figure}

\newpage
\subsubsection*{conv3\_1}.
\begin{figure}[H]
\centering
\includegraphics[width=\textwidth]{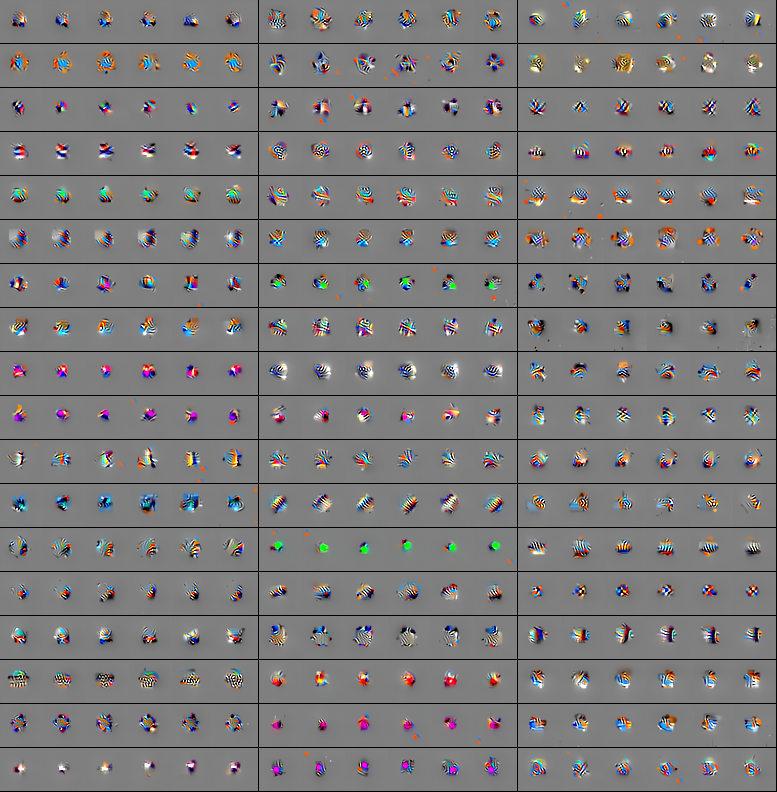}
\end{figure}

\end{document}